\documentclass{article}

\AtBeginDocument{%
  }
    
  \usepackage[eandd, final]{neurips_2026}

\usepackage{graphicx}
\usepackage{booktabs}
\usepackage{array}
\usepackage{listings}
\usepackage{minted}
\usepackage[utf8]{inputenc} 
\usepackage[T1]{fontenc}    
\usepackage{hyperref}       
\usepackage{url}            
\usepackage{booktabs}       
\usepackage{amsfonts}       
\usepackage{nicefrac}       
\usepackage{microtype}      
\usepackage{tabularx} 
\usepackage[table]{xcolor}
\usepackage{xcolor}         
\usepackage{amsmath} 
\usepackage{amssymb} 
\usepackage{multicol}

\definecolor{codegreen}{rgb}{0,0.6,0}
\definecolor{codegray}{rgb}{0.5,0.5,0.5}
\definecolor{codepurple}{rgb}{0.58,0,0.82}
\definecolor{backcolour}{rgb}{0.95,0.95,0.92}

\lstdefinestyle{mystyle}{
    backgroundcolor=\color{backcolour},   
    commentstyle=\color{codegreen},
    keywordstyle=\color{magenta},
    numberstyle=\tiny\color{codegray},
    stringstyle=\color{codepurple},
    basicstyle=\ttfamily\footnotesize,
    breakatwhitespace=false,         
    breaklines=true,                 
    captionpos=b,                    
    keepspaces=true,                 
    numbers=left,                    
    numbersep=5pt,                  
    showspaces=false,                
    showstringspaces=false,
    showtabs=false,                  
    tabsize=2
}

\title{Evaluating Developmental Cognition Capabilities of LLMs}

\author{%
    Xiao Xiao\\
 De Vinci Research Center, France\\
   MIT Media Lab, USA\\
  \And
  Hayoun Noh\\
  University of Oxford, UK\\
  \And
  Mar Gonzalez-Franco\\ 
  Google, USA\\
  \texttt{margon@google.com}\\
}

\begin{document}

\maketitle

\begin{abstract}
Conversational AI is increasingly personalized around users’ preferences, histories, goals, and knowledge, but much less around how users interpret and take up model outputs to construct and understand their reality. We draw on Robert Kegan’s constructive-developmental theory as a complementary lens on this dimension. Existing methods for assessing developmental stage in the Keganian tradition rely either on expert interviews that do not scale or on sentence-completion instruments that are proprietary, lengthy, or invasive. To make this perspective tractable for LLM evaluation, we introduce the \textbf{Developmental Sentence Completion Test (DSCT)}, a 20-item instrument designed to elicit developmental signal in self-administered text. Throughout, we treat the resulting labels as characterizations of stage-like structure in elicited responses, not as validated person-level developmental stage. We then ask how much of that signal can be recovered by LLMs across three elicited response regimes: simulated personas, real human respondents, and default model-generated answers.

On simulated personas, top frontier models recover simulator-intended labels with high accuracy. On real human DSCT responses, human--LLM agreement is fair, with much stronger within-neighborhood than exact agreement. Finally, when LLMs answer DSCT prompts without persona-conditioning, their responses exhibit stable stage-like differences across model families, with larger and newer models tending to generate higher-rated text. These results suggest that stage-conditioned signal is cleaner in synthetic responses than in human-written DSCT text, and that the core constraint for stage-aware conversational AI is not classifier accuracy alone, but the availability of developmental signal from elicited text.
\end{abstract}

\section{Introduction}

Conversational AI systems are increasingly used to support learning, reasoning, and decision-making, including in educational, reflective, and counseling-oriented settings \citep{wang2026large, li2025system, chiu2024computational}. Personalization has therefore become an active research direction, typically framed around users' preferences, histories, contexts, goals, or knowledge states \citep{chen2024large}. Such approaches mostly adapt to what users want, know, or are trying to accomplish, and much less to how they interpret and take up model outputs. Developmental psychology, however, suggests that people also differ in how they make sense of experience and knowledge \citep{piaget1952origins, kegan1994}.

One useful lens on these differences comes from Kegan’s constructive-developmental theory, which characterizes development in terms of subject--object transformation: changes in what a person is subject to and cannot yet step back from, versus what they can take as an object of reflection and evaluate \citep{kegan1994}. In this work, we use Kegan’s framework to ask to what extent differences in meaning-making can be recovered from elicited text. This matters for conversational AI because the same model output may function differently for users whose expressed meaning-making structure differs, even when their stated goals are similar. Ignoring this variation may contribute to familiar failure modes such as homogenized outputs \citep{jiang2025artificial} or sycophantic reinforcement of user beliefs \citep{chandra2025sycophantic}. Note that our goal is not to measure a user’s developmental stage directly, but to test whether elicited text contains recoverable clues about meaning-making structure.

Standard methods for assessing developmental structure in the Keganian tradition are not well suited to LLM benchmarking. The best-known approach, the Subject--Object Interview and related developmental interviews, provides rich access to meaning-making structure but requires extended dialogue and expert interpretive coding \citep{laske2023process, kegan1994}. While Kegan’s canonical assessments are interview-based, sentence-completion tests have emerged as another way to asses meaning-making structure \citep{loevinger1998technical, cook1999postautonomous}. Here, respondents complete short open-ended stems, and the resulting text is assessed structure of meaning-making it expresses. This format is especially relevant for LLM evaluation. As LLMs are both producers and readers of text, they can generate sentence-completion responses and classify such responses from either humans or other models. However, sentence-completion instruments such as Loevinger’s and Cook-Greuter’s
instruments are not designed to assess Keganian developmental structure directly, and they are often proprietary, lengthy, or include gendered and personally invasive prompts \citep{loevinger1998technical, cook1999postautonomous}. We therefore introduce the Developmental Sentence Completion Test (DSCT), a 20-item self-administered instrument designed to elicit text rich enough for trained human raters or LLMs to assign a tentative Keganian label to a response set, while removing the proprietary, gendered, and invasive items of older instruments.

Our central question is: how much developmental signal can be recovered from DSCT-style text, and how does that depend on who or what produced the text? To answer this, we compare three regimes: simulated personas, real human respondents, and default model-generated answers. First, we study synthetic DSCT responses generated from expert-described developmental profiles, asking whether stage-conditioned signal can be recovered by LLM classifiers and corroborated by trained human raters. Second, we apply DSCT to human respondents and measure agreement between human raters and LLM classifiers on the resulting response sets. Third, we prompt LLMs to answer DSCT items without persona-conditioning and analyze the developmental structure of the text they produce. These three regimes let us compare human-written, model-simulated, and model-generated responses within a shared elicitation format.

Our contributions are threefold. First, we introduce DSCT, a 20-item sentence-completion instrument for matched human and LLM evaluation of stage-like structure in elicited text. Second, we benchmark twelve LLMs on Keganian labeling of simulated and human DSCT responses against trained human raters, showing that top frontier models recover simulator-intended labels with high accuracy under controlled synthetic conditions, while developmental signal is less clean but still sufficiently recoverable in human-written DSCT responses for substantial agreement on broader stage regions. Third, we analyze default DSCT responses generated by LLMs, showing that larger and newer models tend to produce text rated at higher developmental stages.

\section{Designing the DSCT}
\label{sec:dsct-design}
The Developmental Sentence Completion Test (DSCT) is a 20-item self-administered instrument designed to elicit text samples sufficient for an LLM or trained human rater to make a tentative assessment of a respondent's likely Developmental Cognition Kegan stage. We describe the design here so its scope and limits are clear before the experiments.

\textbf{Scope and limitations.} The Subject--Object Interview \cite{kegan1994} remains the gold standard for stage assessment, but its 60--90 minute semi-structured format and certified-coder requirement are incompatible with computational evaluation at scale. Sentence-completion tests --- the Loevinger SCT \cite{loevinger1998technical} and Cook-Greuter's MAP / SCTi-MAP \cite{cook1999postautonomous} are the canonical instruments --- offer a lighter-weight alternative: respondents complete short stems in their own words, and structural features of the response are used to infer meaning-making complexity. DSCT inherits the SCT/MAP format and is intended to support the empirical question of \emph{how much developmental signal can be recovered from a brief, self-administered text sample}. However, the DSCT should not be considered a full substitute for the SOI or SCTi-MAP, neither a diagnostic instrument, and it is not validated for individual high-stakes decisions. The DSCT is also not a measure of intelligence, education, or verbal ability, although responses unavoidably co-vary with these. ``Stage,'' as we use the term, is a property of how a \emph{response} is structured, not a stable trait we attribute to the respondent.

\textbf{Item construction.} We started from the affective territory probed by the SOI, which uses prompt cards covering recurrent meaning-making situations: \emph{angry / mad}, \emph{torn / conflicted}, \emph{sad}, \emph{success}, \emph{strong stand / conviction}, \emph{moved / touched}, \emph{loss / farewells}, \emph{change}, \emph{important}, and \emph{anxious / nervous}. For each territory we generated candidate vignettes with LLM assistance and then curated them by hand: two of the authors independently selected the vignettes they judged structurally richest and least culturally loaded, retaining only items both endorsed. This yielded two vignettes per territory, 20 stems total. The two vignettes per territory are written in different voices: one in the first person (Section 1, \emph{self-assessment}, items 1--10) and one in the third person about a generic other (Section 2, \emph{abstracted-other assessment}, items 11--20). The two-voice design accommodates respondents who default to socially desirable first-person responses but reveal more structural complexity when reasoning about another person in the same situation; the two sections probe the same constructs through different framings rather than measuring separate dimensions. 

\textbf{Comparison to SCT.} 
Compared to the 36-item Loevinger SCT, DSCT is 44\% shorter and removes items targeting gender roles, sexuality, and family relationships (e.g., \emph{``A man's job\ldots''}, \emph{``Usually she/he felt that sex\ldots''}, \emph{``A wife/husband should\ldots''}) which we judged invasive in a self-administered online setting and orthogonal to structural complexity in meaning-making. We piloted the resulting instrument informally on ourselves before the experiments. Full item lists for both questionnaires are in Appendix A.1.

It is worth noting that despite the DSCT improvements compared to the SCT with respect to cultural norms, it is still not fully culture-free: in the current version stems are in English, and the underlying frame reflects Western adult-development scholarship.

\subsection{Experiment 1: Controlled validation on simulated personas}

We begin with a controlled validation setting. Because no large-scale corpus of stage-labeled DSCT responses exists, we use simulated personas anchored in 23 expert-described developmental profiles drawn from prior literature \citep{bartone2002cognitive,berger2024changing,laske2023process,baxtermagolda2007interview} (the 23 profiles used for simulating the personas is available in the Appendix~\ref{app:agent-profiles}). The purpose of this experiment is not to establish real-world stage inference, but to test whether stage-conditioned developmental signal embedded in synthetic DSCT responses can be recovered by LLM classifiers, and to what extent those simulator-intended labels are corroborated by trained human raters. For further validation an additional comparison between the DSCT and the longer 36-item Loevinger SCT, run on a subset of classifiers, is reported in Appendix~\ref{app:dsct-vs-sct}.

\textbf{Simulating the personas.} Each of the 23 profiles specifies a target stage (solid stages 2--5, plus transitional 2/3, 3/4, and 4/5) together with a brief description of the corresponding meaning-making structure from the literature. We used Gemini 3.1 Pro to generate DSCT responses conditioned on each profile, prompting it to adopt the persona’s worldview and tone while avoiding overt lexical cues that would reveal the target stage too directly (full simulator prompt and system instruction in Appendix~\ref{app:exp1-prompts}). To account for stochasticity, we generated six independent responses per profile, yielding 138 simulated cases. This is a best practice in AI benchmarking to remove single-run ranking inversions \citep{alvarado2025repetitions}, -- 3 iterations remove over ~83\% of stochastic effects. Note that even with multiple simulations, these cases still only instantiate simulator-intended stage structure in text; they do not constitute human ground truth. The human-rating step below therefore serves as a check on how well the intended structure is actually realized in the generated responses.

\textbf{Human rating of a stratified subset.} To assess whether the simulator-intended targets were realized in a form that trained readers would endorse, we sampled 46 simulated cases, corresponding to two independently generated response sets for each of the 23 profiles, and had two raters with basic orientation to Kegan’s constructive-developmental theory, supported by a brief rating guide, evaluate them independently. Raters saw only the DSCT responses with random IDs and were blind to the target stage, profile descriptions, and one another’s ratings. Inter-rater reliability was high (quadratic-weighted $\kappa = 0.927$), and agreement between rater consensus and simulator-intended stage was 65.2\% exact and 100\% within $\pm 0.5$ stage, indicating that even when raters diverged from the intended label, they never missed by more than a half-stage. We use this comparison not as validation of a human ground truth, but as a check on whether the generated responses exhibit the intended developmental structure in a form that human raters can recognize. It also allows us to compare LLM judgments not only to simulator-intended labels, but also to human ratings of the same responses. Full protocol details are reported in Appendix~\ref{app:rater-protocol}.

\textbf{LLM classification across twelve models.} For each simulated case, we asked twelve LLMs spanning major frontier model families (Claude Opus 4.6, Claude 4.5 Haiku, GPT 5.5, GPT 5 Mini, Grok 4.2, DeepSeek V4, DeepSeek R1, Gemini 3.1 Pro, Gemini 3.1 Flash, Mistral 3 Large, Qwen 3.6 Plus, Kimi K2.6) to classify the response into one of stages 1--5 or a transitional stage. The classifier prompt instructed the model to act as a developmental psychologist familiar with Kegan’s theory and to provide a brief rationale alongside each stage assignment (full prompt in Appendix~\ref{app:exp1-prompts}). Because Gemini 3.1 Pro is both the simulator and one of the classifiers, its results are reported for completeness but should be interpreted with particular caution.

\textbf{Results.}

\begin{figure}[t]
    \centering
    \includegraphics[width=0.9\linewidth]{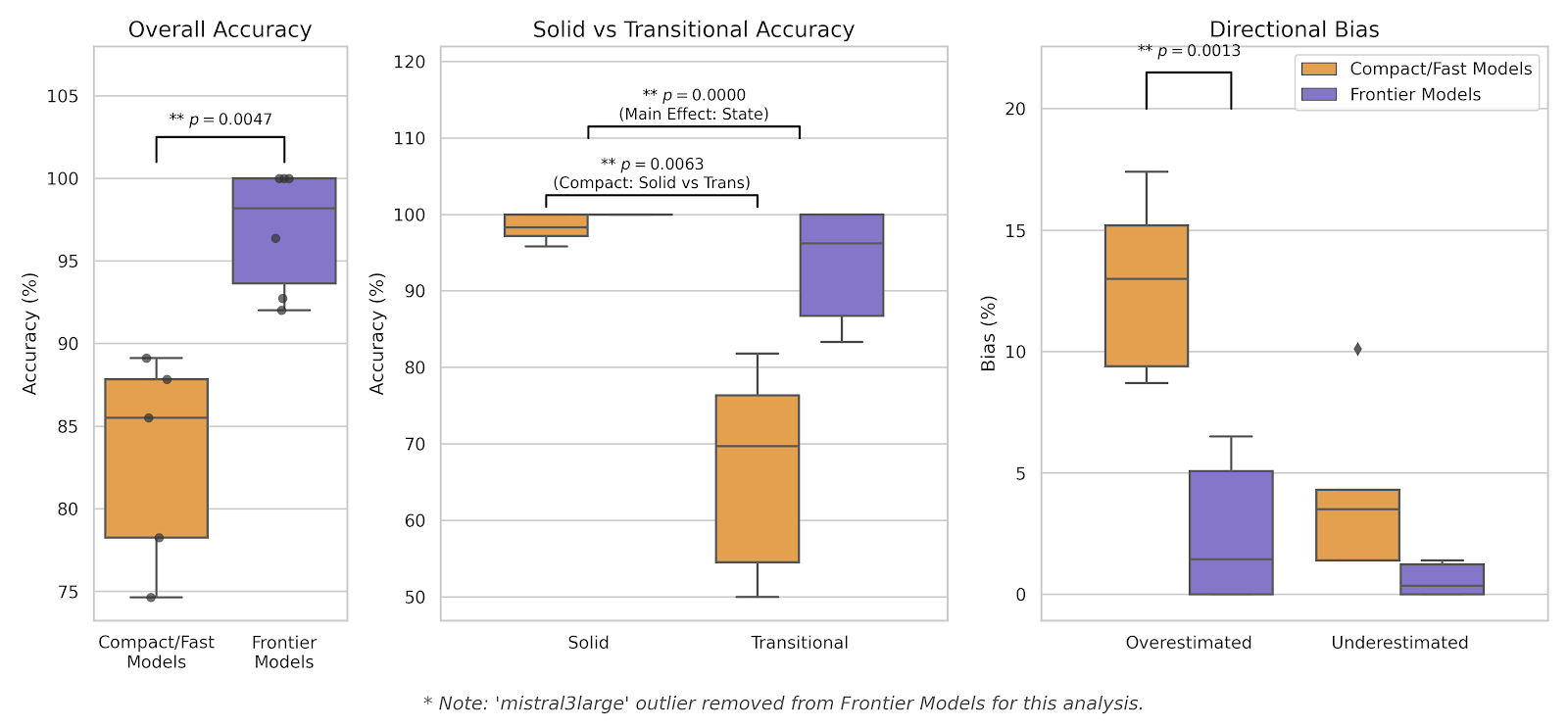}
    \caption{\textbf{Experiment 1 summary on simulated personas.} Left: overall accuracy differs by model tier, with frontier models outperforming compact/fast models. Middle: the largest performance drop occurs on transitional stages, especially for compact/fast models. Right: errors are directionally asymmetric, with compact/fast models showing stronger upward bias than frontier models.}
    \label{fig:exp1_results}
\end{figure}

\textit{Per-model classification accuracy.} Table~\ref{model-performance-table} summarizes per-model performance against simulator-intended target stages, ordered by Cohen's Kappa. Three models --- Claude Opus 4.6, Gemini 3.1 Pro, and Grok 4.2 --- recovered the intended stage label on all simulated cases. Five further models (DeepSeek V4, GPT 5.5, DeepSeek R1, Kimi K2.6, GPT 5 Mini) recovered all solid-stage cases but degraded on transitional cases, where the structural distinction between, e.g., a Stage 3/4 and a Stage 4 response is finer-grained. Claude 4.5 Haiku, Qwen 3.6 Plus, Gemini 3.1 Flash, and Mistral 3 Large performed less consistently across both categories.

\begin{table}[htbp!]
  \tiny
   \caption{Average performance metrics across various models and human raters. Ordered by quadratic-weighted Cohen's $\kappa$. \textbf{Note:} all model rows are computed on the full 138 simulated cases; the Human Raters row is computed on the 46-case stratified subset rated by two human judges.}
   \label{model-performance-table}
   \centering
   \begin{tabular}{l c c c c c c}
      \toprule
      Model Type & \shortstack{Avg Kappa $\kappa$ \\ ($\pm$SD)} & \shortstack{Accuracy \\ ($\pm$SD)} & \shortstack{Solid Stage \\ Accuracy ($\pm$SD)} & \shortstack{Transitional Stage \\ Accuracy ($\pm$SD)} & \shortstack{Bias \% (Over/ \\ Underestimate)} & \shortstack{Avg. Offset \\ Magnitude (+/-)} \\
      \midrule
      Claude Opus 4.6 & 1.0 & 100.0\% & 100.0\% & 100.0\% & +0.0 / -0.0 \\
      Gemini 3.1 Pro & 1.0 & 100.0\% & 100.0\% & 100.0\% & +0.0 / -0.0 \\
      Grok 4.2 & 1.0 & 100.0\% & 100.0\% & 100.0\% & +0.0 / -0.0 \\
      DeepSeek V4 & 0.95 $\pm$ 0.11 & 96.4\% $\pm$ 8.9 & 100.0\% & 92.4\% $\pm$ 18.6 & \textcolor{orange}{+2.9\%} / \textcolor{blue}{-0.7\%} & \textcolor{orange}{+0.50} / \textcolor{blue}{-0.50} \\
      \rowcolor{gray!20} Human Raters (46-case subset) & 0.927 & 65.2\% $\pm$ 24.7 & 91.3\% $\pm$ 28.1 & 39.1\% $\pm$ 25.3 & \textcolor{orange}{+28.3\%} / \textcolor{blue}{-6.5\%} & \textcolor{orange}{+0.50} / \textcolor{blue}{-0.50} \\
      GPT 5.5 & 0.91 $\pm$ 0.12 & 92.8\% $\pm$ 9.8 & 100.0\% & 84.8\% $\pm$ 20.5 & \textcolor{orange}{+5.8\%} / \textcolor{blue}{-1.4\%} & \textcolor{orange}{+0.50} / \textcolor{blue}{-0.50} \\
      DeepSeek R1 & 0.901 $\pm$ 0.10 & 92.0\% $\pm$ 8.0 & 100.0\% & 83.3\% $\pm$ 16.7 & \textcolor{orange}{+6.5\%} / \textcolor{blue}{-1.4\%} & \textcolor{orange}{+0.50} / \textcolor{blue}{-0.50} \\
      Claude 4.5 Haiku & 0.86 $\pm$ 0.11 & 89.1\% $\pm$ 9.0 & 95.8\% $\pm$ 7.0 & 81.8\% $\pm$ 22.3 & \textcolor{orange}{+9.4\%} / \textcolor{blue}{-1.4\%} & \textcolor{orange}{+0.54} / \textcolor{blue}{-0.50} \\
      Qwen 3.6 Plus & 0.85 $\pm$ 0.16 & 87.8\% $\pm$ 12.8 & 98.3\% $\pm$ 3.7 & 76.4\% $\pm$ 26.2 & \textcolor{orange}{+8.7\%} / \textcolor{blue}{-3.5\%} & \textcolor{orange}{+0.50} / \textcolor{blue}{-0.50} \\
      Kimi K2.6 & 0.82 $\pm$ 0.15 & 85.5\% $\pm$ 12.2 & 100.0\% & 69.7\% $\pm$ 25.5 & \textcolor{orange}{+13.0\%} / \textcolor{blue}{-1.4\%} & \textcolor{orange}{+0.50} / \textcolor{blue}{-0.50} \\
      GPT 5 Mini & 0.73 $\pm$ 0.0 & 78.3\% & 100.0\% & 54.5\% & \textcolor{orange}{+17.4\%} / \textcolor{blue}{-4.3\%} & \textcolor{orange}{+0.50} / \textcolor{blue}{-0.50} \\
      Gemini 3.1 Flash & 0.69 $\pm$ 0.05 & 74.6\% $\pm$ 4.3 & 97.2\% $\pm$ 4.3 & 50.0\% $\pm$ 9.5 & \textcolor{orange}{+15.2\%} / \textcolor{blue}{-10.1\%} & \textcolor{orange}{+0.50} / \textcolor{blue}{-0.57} \\
      Mistral 3 Large & 0.38 $\pm$ 0.28 & 48.6\% $\pm$ 24.0 & 73.6\% $\pm$ 15.3 & 21.2\% $\pm$ 39.3 & \textcolor{orange}{+12.3\%} / \textcolor{blue}{-39.1\%} & \textcolor{orange}{+0.50} / \textcolor{blue}{-0.97} \\
      \bottomrule
   \end{tabular}
\end{table}

\textit{Frontier vs.\ compact models.} Aggregating across model tiers (Compact/Fast: Haiku, GPT 5 Mini, Flash; Frontier: the rest), a two-way ANOVA on accuracy with factors Model Tier and Stage Type (Solid vs.\ Transitional) showed a main effect of Stage Type ($p < 0.0001$ with Mistral excluded as outlier; $p < 0.001$ with Mistral included). compact/fast models maintained $98.3\%$ accuracy on solid stages but dropped to $66.5\%$ on transitional stages ($t=5.12$, $p=0.006$, Welch's). Full statistics, including with-outlier and without-outlier analyses, are in Appendix~\ref{app:exp1-stats}.

\textit{Directional bias differs by model tier.} Among the three top frontier models, classification was effectively unbiased on the simulated set: Claude Opus 4.6, Gemini 3.1 Pro, and Grok 4.2 each produced zero overestimation and zero underestimation. The remaining frontier models showed small overestimation effects on transitional cases only ($+2.9\%$ to $+6.5\%$). By contrast, compact/fast models systematically overestimated stages ($M=12.7\%$, $SD=3.3$), an effect significantly larger than the frontier tier ($M=1.8\%$, $SD=2.9$; $t=6.07$, $p=0.001$, Welch's). Mistral 3 Large was the sole model to underestimate, doing so substantially ($-39.1\%$ on transitional cases). Under these controlled synthetic conditions, smaller models therefore fail in a specific direction, tending to assign higher stages rather than producing random errors.

\textbf{Takeaways.} (i) Under controlled synthetic conditions, top frontier LLMs recover simulator-intended stage labels with high accuracy. (ii) Smaller models degrade especially on transitional cases, where the structural distinction is finer, and their errors are more often upward than random. (iii) Agreement with trained human raters was lower than agreement with simulator-intended labels, indicating that the synthetic responses contain signal that is more easily recoverable relative to the simulator target than to human raters. Experiment 1 should therefore be read as a controlled validation of recoverable developmental signal in synthetic DSCT responses, not as evidence that stage can be inferred equivalently well from real human language. This experiment provides additional evidence that the newly proposed DSCT is a sensitive metric for Kegan stage evaluation.

\section{Experiment 2: DSCT on human participants}

Experiment 1 establishes that frontier LLM classifiers recover simulator-intended target stages on synthetic personas under controlled conditions. The harder question is whether the same instrument and the same classifiers behave sensibly on real human respondents, where structural signal is noisier, response length and engagement vary, and there is no target stage to recover. We address this by collecting DSCT responses from human participants and using ratings from three Kegan-trained raters as the human reference label.

\textbf{Participants.} We recruited 83 participants via email lists. The sample consisted of 49 males (59.0\%) and 34 females (41.0\%), with an average age of 32.04 years ($SD = 10.96$; range = 20--64 years). Participants represented 19 countries, with the largest groups residing in France (41.5\%), the United States (17.1\%), Spain (9.8\%), and South Korea (6.1\%). The cohort was highly experienced with Generative AI technologies: 79.7\% of participants reported using Large Language Model (LLM) chatbots daily, and an additional 13.9\% used them more than twice a week. When asked for their primary AI chatbot of choice, 50.0\% reported OpenAI’s ChatGPT, followed by Anthropic’s Claude (23.8\%) and Google’s Gemini (16.2\%). Recruitment was through a sign-up form that explained the study purpose, voluntary participation, and the right to withdraw at any time. The study did not require formal IRB review under the institution's policies for low-risk online survey research; the consent procedure and full sign-up text are reproduced in Appendix~\ref{app:user-study}, and the experimenters followed the Helsinki declaration guidelines.

\textbf{Task.} Participants completed the DSCT, presented as a \emph{Situational Sense-Making Questionnaire} to avoid priming the developmental construct. They were also invited to complete a separate exploratory follow-up task involving their regular LLM chatbot, described in Appendix~\ref{app:chat-followup}.

\textbf{Human rating.} Each participant’s 20-item DSCT response set was rated independently by three raters with basic orientation to Kegan’s constructive-developmental theory, supported by a brief rating guide, following the same protocol as in Experiment 1: blind independent rating, followed by aggregation into a downstream consensus label (full protocol in Appendix~\ref{app:rater-protocol}). Overall agreement across the three raters was fair (Fleiss' $\kappa = 0.31$). Pairwise quadratic-weighted Cohen's $\kappa$ ranged from 0.49 to 0.81, indicating that some rater pairs agreed substantially more than others. Of the 83 cases, 23 received unanimous ratings, 52 received a majority rating (2 of 3 raters agreeing), and 8 produced full three-way disagreement. We therefore use the term \emph{rater consensus} rather than \emph{ground truth} throughout: stage labels derived from a 20-item self-administered instrument are limited compared to the Subject--Object Interview, and the rater consensus is the best human reference label available at this scale rather than a definitive measurement.

\textbf{Sample distribution.} Using majority vote wherever at least two raters agreed, 75 of 83 participants received a usable consensus label, while 8 remained unresolved due to full disagreement. The resulting distribution was: Stage 2/3 ($n=3$), Stage 3 ($n=18$), Stage 3/4 ($n=17$), Stage 4 ($n=28$), and Stage 4/5 ($n=9$). The sample contains no consensus cases at Stage 2 or Stage 5, consistent with population estimates for these stages \citep{kegan1994} and with a self-selected sample drawn from professional and academic networks. We treat this as a meaningful scope limitation: claims in this section apply primarily to the Stage 3 / 3-4 / 4 / 4-5 range, where most adult variation occurs, and not to the developmental extremes.

\textbf{LLM classification of human responses.} We then asked two of the top performing LLMs from Experiment 1 (Claude Opus 4.6 and Gemini 3.1 Pro) to classify each participant's DSCT response set, using the identical classifier prompt (Appendix~\ref{app:exp1-prompts}). For each participant this produces three quantities of interest: (i) the rater-consensus stage, (ii) the per-classifier stage from each of the top models, and (iii) the cross-classifier consensus stage (computed as the modal label across the models, with ties broken by averaging numeric stage values where applicable; ties and ambiguous cases are reported separately). Comparing these allows us to ask whether classifier judgments on real human DSCT data agree with trained human raters, and whether the agreement pattern matches what we observed on simulated personas.

\begin{figure}[h!]
    \centering
    \begin{minipage}[t]{0.5\linewidth}
        \centering
        \includegraphics[width=\linewidth]{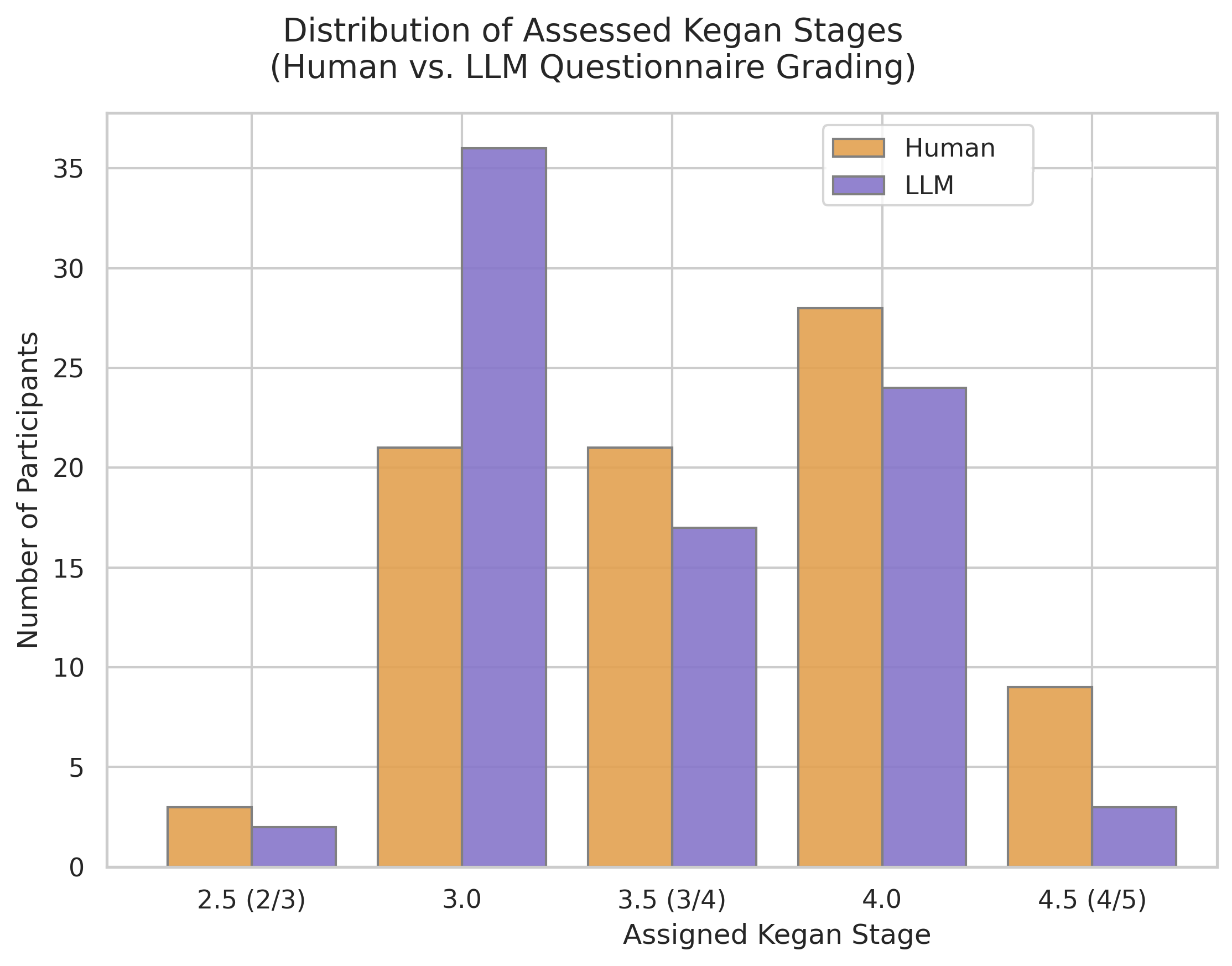}
    \end{minipage}\hfill
    \begin{minipage}[t]{0.45\linewidth}
        \centering
        \includegraphics[width=\linewidth]{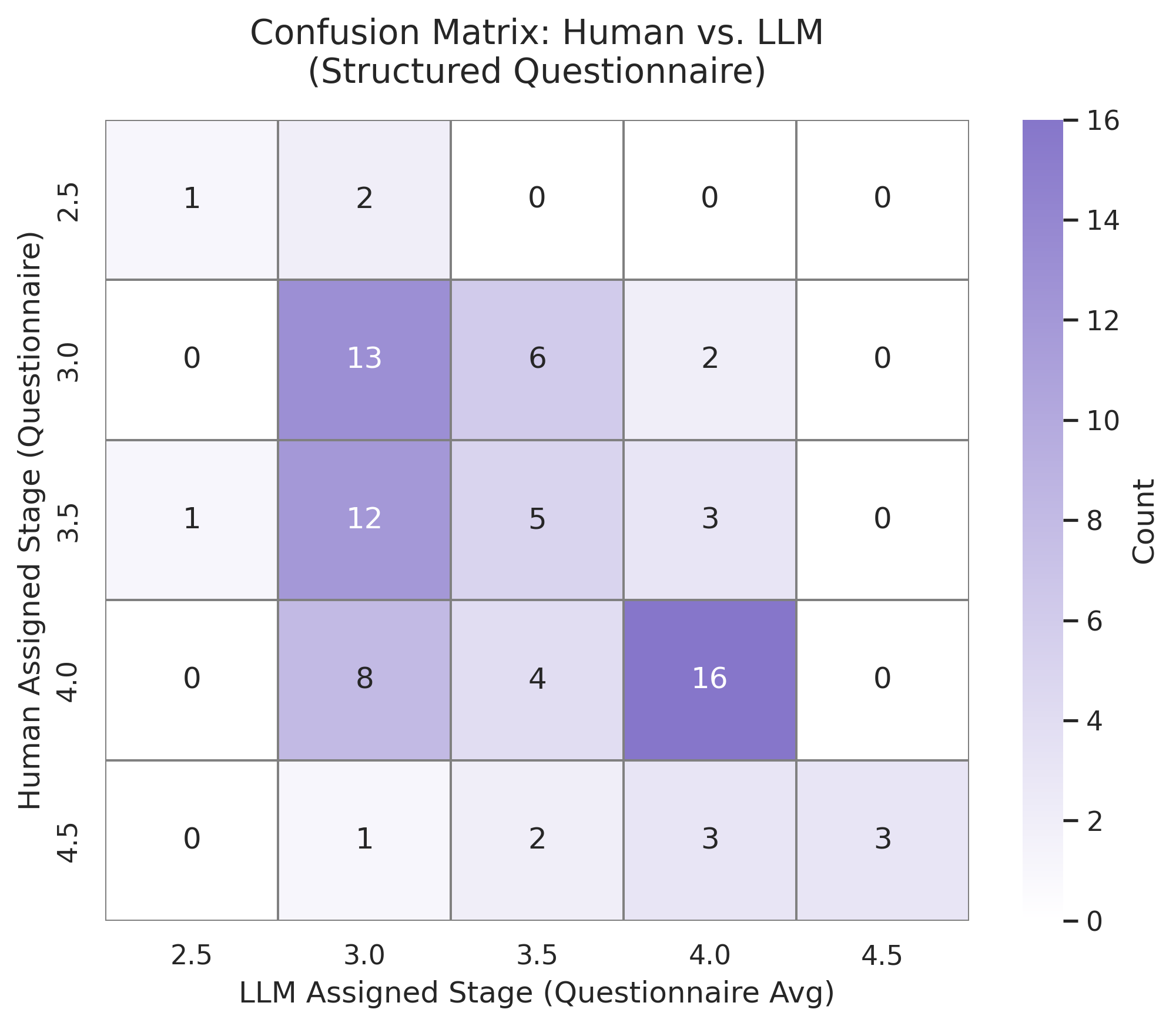}
    \end{minipage}
    \caption{\textbf{Experiment 2: human vs.\ LLM agreement on structured DSCT responses.} Left: marginal stage distributions assigned by human and LLM raters. Right: agreement heatmap. Most disagreements remain near the diagonal, indicating that LLM judgments often fall within the participant's developmental neighborhood even when exact agreement is limited.}
    \label{fig:exp2_questionnaire}
\end{figure}

\textbf{Results.} \textit{Human--LLM agreement on structured DSCT.} On the questionnaire data (N=83), agreement between the human reference labels and the LLM evaluator was fair (quadratic weighted $\kappa = 0.49$). Exact agreement was 46.3\% ($n=38$), and agreement within $\pm 0.5$ stage was 82.9\% ($n=68$). The human rater assigned a mean stage of 3.62 ($SD=0.53$), while the LLM assigned a mean stage of 3.44 ($SD=0.53$); this difference was statistically significant (Wilcoxon signed-rank, $W=244.5$, $p=0.002$). Thus, on structured DSCT responses from real humans, the LLM generally locates the participant's developmental neighborhood, but does not reliably reproduce exact human staging. 

\textit{Directional asymmetry.} Disagreements were not symmetric. In 37.8\% of all cases ($n=31$), the human evaluator assigned a higher stage than the LLM, whereas the LLM assigned a higher stage in only 15.9\% of cases ($n=13$). On structured questionnaire input, the LLM therefore appears slightly more conservative than the human rater rather than systematically inflationary (Figure~\ref{fig:exp2_questionnaire}). 

An exploratory follow-up using participants' unstructured chat history also produced strongly upward-skewed ratings and very weak agreement with structured questionnaire-based judgments, suggesting that stage inference from unconstrained chat is not yet reliable in the present setting (Appendix~\ref{app:chat-followup}).

\textbf{Interpretation.} This pattern contrasts with the cleaner synthetic performance of Experiment~1. On real human DSCT responses, stage-relevant signal remains recoverable, but the ceiling is lower because the elicited text is less internally consistent and provides a less uniform target for both human raters and LLM classifiers. This is reflected not only in lower model agreement than in the synthetic setting, but also in the human labels themselves: while most participants received a usable consensus label, a non-trivial minority remained ambiguous even for trained raters. Additional item-level analyses supported this interpretation, suggesting that predictive signal in the human data was concentrated in a subset of questions rather than expressed uniformly across the questionnaire (Appendix~\ref{app:exp2-random-forest}). The benchmark target in this setting is therefore not exact stage recovery, but approximate alignment with trained human judgment on elicited text.

\textbf{Takeaways.} (i) DSCT applied to real human respondents yields usable but imperfect developmental labels: agreement is fair, exact matches are limited, and proximity-based agreement is substantially higher than exact agreement. (ii) On structured questionnaire text, the LLM is slightly stricter than the human rater rather than systematically upward-biased. (iii) Compared to Experiment~1, performance drops on real human responses, indicating that recoverability depends not only on the classifier, but also on the source and cleanliness of the elicited text.

\subsection{Experiment 3: Stage-like structure in default model-generated DSCT responses}

Experiments 1 and 2 ask whether developmental signal in DSCT responses can be recovered from synthetic personas and human respondents. A complementary question is whether models also differ in the developmental structure of the text they generate by default. To examine this, we prompted a broader set of language models to answer the DSCT without persona-conditioning or target-stage instructions, and then rated the resulting responses using the same developmental framework.

\textbf{Design.} We grew from our original cohort of 12 frontier models and prompted 29 language models, ranging from GPT-3.5 Turbo to current frontier systems, to complete the full 20-item DSCT in a single pass. The prompt presented the questionnaire exactly as a respondent would see it, instructed the model to answer every item, and requested JSON-formatted responses (prompt in Appendix~\ref{app:exp3-prompts}). Unlike Experiment 1, no developmental profile or target stage was supplied. The resulting outputs therefore reflect the model's default response style under DSCT elicitation rather than stage-conditioned simulation.

\textbf{Model set.} The 29 models included 18 frontier models and 11 compact/fast models, spanning releases from early 2023 through the present. This allows us to ask not only whether models differ in the stage-like structure of their outputs, but also whether that structure changes over model generations and between model tiers.

\textbf{Rating procedure.} The DSCT responses generated by each model were rated by Gemini 3.1 Pro using the same Keganian rubric as in Experiments~1 and~2. To reduce single-run stochasticity, we repeated the rating procedure three times per model and used the resulting labels jointly in analysis, following recent recommendations for repeated evaluation in LLM benchmarking \citep{alvarado2025repetitions}. No human raters were involved in this experiment. We interpret the resulting labels not as claims that LLMs possess human developmental stages, but as ratings of the stage-like structure of the text they generate under a shared elicitation format.

\begin{figure}[h!]
    \centering
    \includegraphics[width=0.8\linewidth]{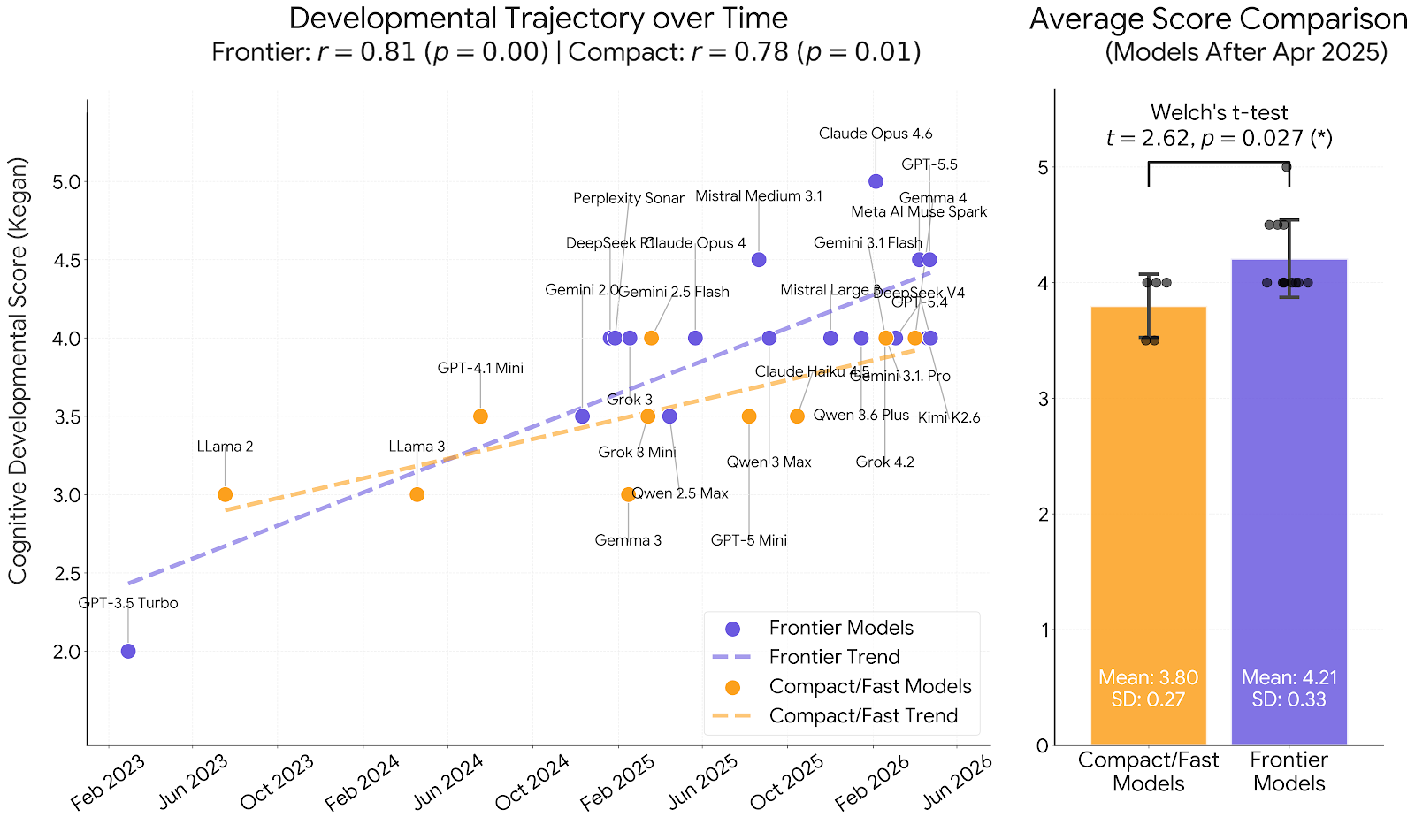}
    \caption{\textbf{Experiment 3: stage-like structure in default model-generated DSCT responses.} Left: developmental ratings of model-generated DSCT responses increase over release time in both frontier and compact/fast model families. Right: among models released after April 2025, frontier models produce higher-rated responses on average than compact/fast models.}
    \label{fig:exp3_results}
\end{figure}

\textbf{Results.} Most current frontier models were rated in the Stage~4 to Stage~4/5 range, with a smaller number reaching Stage~5-like structure. compact/fast models were rated lower on average, with several clustering around Stage~3/4 and one model (GPT-3.5 Turbo) rated at Stage~2 (Figure \ref{fig:exp3_results}). Across the full sample, frontier models exhibited a higher mean developmental score ($M=3.97$, $SD=0.61$) than compact/fast models ($M=3.55$, $SD=0.42$). Full per-model ratings are reported in Appendix~\ref{app:exp3-model-table}.

We also observe a temporal trend: newer models tend to generate DSCT responses rated at higher stages than older ones. This pattern is visible in both frontier and compact/fast families. Across the full set of models, developmental score increased with release date, and the overall regression was significant ($F(3,25)=19.22$, $p<0.001$, $R^2=0.698$), although the interaction between release time and model tier did not reach significance. We therefore interpret the main pattern as a general upward shift in stage-like output structure over successive model generations rather than a sharply diverging slope between tiers.

\textbf{Interpretation.} These findings suggest that developmental structure matters not only on the classifier side, but also in the default style of model-generated text. Models that generate more self-authored or self-transforming DSCT responses may also be more likely to interpret others through those same structures. This provides one plausible mechanism for systematic upward pressure in stage assignment: a model whose own outputs default to Stage~4 or Stage~4/5 language may also tend to read elicited responses through that lens.

At the same time, these ratings should be interpreted cautiously. We do not claim that LLMs possess human developmental stages, nor that stage labels capture stable internal properties of a model. Rather, Keganian ratings provide a descriptive framework for comparing the structure of model-generated responses under a shared elicitation format.

\textbf{Takeaways.} (i) LLMs differ not only in their ability to classify developmental structure, but also in the stage-like structure of the DSCT responses they generate by default. (ii) Larger and newer models tend to produce text rated at higher developmental stages than smaller or older models. (iii) Model-side output structure may therefore influence not only what models say, but also how they read stage-like structure in the text of others.

\section{Discussion}
\label{sec:discussion}

\paragraph{Recoverable developmental signal in text.}
The main result of this paper is that DSCT makes developmental signal recoverable in text, but not equally across sources. Under stage-conditioned simulation, frontier models recover intended labels with high accuracy. On real human DSCT responses, agreement is weaker and bounded by the fact that even trained raters do not fully agree. Simulated personas produce cleaner stage-conditioned text than human respondents, while human DSCT responses are noisier, shorter, and more uneven. This establishes a useful but limited benchmark target: not developmental structure in the person, but recoverable developmental signal in elicited text.

\paragraph{Model-side developmental style.}
A notable finding is that models differ not only in how they classify developmental structure, but also in the structure of the responses they generate under the same DSCT prompt. Larger and newer models tend to produce outputs rated at higher stages than smaller and older ones. We do not interpret this as evidence that LLMs possess developmental stages in the human sense. Rather, Keganian ratings provide a descriptive lens on output structure. One possible implication is that models whose default outputs are more self-authored or system-level may also be more likely to interpret others through that same frame, though this relationship remains to be tested directly.

\paragraph{Developmental modeling and the Stage~3/4 region.}
The most consequential region for conversational AI remains the space around Stage~3, Stage~4, and their transition. Stage~3-like organization is more shaped by external recognition, belonging, and relational alignment. Stage~4-like organization supports more independent evaluation from an internal frame \citep{kegan1994,kegan1998over}. This distinction matters because the same model behavior may function differently depending on the meaning-making structure expressed in a user’s elicited text. The point is not that users must reach a particular stage before safely engaging AI as suggested by \citep{aruleswaran6114166emergence}, but that developmental variation may matter for how model outputs are interpreted and taken up. More broadly, this benchmark is not a test of theory of mind in the classical sense \citep{frith2005theory}: what is being compared across all three regimes is not hidden mental state, but recoverable structure in elicited language.

\paragraph{Implications and limits.}
The immediate implication is not that systems should infer developmental stage from arbitrary interaction traces, but that structured elicitation may provide a more reliable basis for developmental adaptation than unconstrained inference. Our exploratory chat-history follow-up reinforces this point: stage assignments derived from unstructured conversational traces were strongly skewed upward and showed very weak agreement with questionnaire-based judgments. DSCT is one possible instrument for structured elicitation of this kind. More broadly, this suggests a complementary direction for personalization: not only adapting to preferences, goals, expertise, or trust-related variables \citep{chen2024large,srinivasan2025adjust}, but also to differences in how users make sense of and take up model outputs. At the same time, DSCT is a brief self-administered instrument and does not substitute for the Subject--Object Interview. Stage labels at this scale should be treated as probabilistic rather than diagnostic. Human agreement is only fair overall, the current sample is concentrated in the Stage~3 to Stage~4/5 range, and Kegan’s framework is only one developmental lens. Future work could compare alternative developmental frameworks, examine state dependence, and test whether similar signal can be recovered under other elicitation formats.

\section{Conclusion}
We introduce DSCT, a 20-item instrument for eliciting developmental signal in self-administered text, and use it to benchmark LLMs on Keganian labeling across three response regimes: simulated personas, real human respondents, and default model-generated answers. Our results show that developmental signal is recoverable from DSCT responses, but not equally across sources. Structured synthetic responses support high classifier performance, real human responses are noisier and bounded by only fair human agreement, and model-generated DSCT responses reveal stable differences in stage-like output structure across model families.

These findings suggest that the central constraint for stage-aware conversational AI is not simply classifier accuracy, but whether elicited text contains developmental structure that can be reliably recovered. If developmental context is to inform interaction, structured elicitation may provide a more reliable basis than passive inference from arbitrary language.

\bibliographystyle{abbrvnat}
\bibliography{references}


\appendix
\section{Appendix}
\subsection{Summary of Kegan stages}
\label{app:kegan-stages}

For orientation, Robert Kegan’s constructive-developmental theory describes a sequence of increasingly complex meaning-making structures. 
\textbf{Stage 1 (Impulsive Mind)} is organized by immediate perceptions and impulses, with little stable reflective distance; this stage is generally rare in adult populations outside severe cognitive impairment. 
\textbf{Stage 2 (Imperialist Mind)} is organized around personal needs, concrete interests, and transactional outcomes; external rules are experienced mainly as constraints or tools. 
\textbf{Stage 3 (Socialized Mind)} is organized through external validation, belonging, and alignment with important others; identity is shaped by relationships and social expectations. 
\textbf{Stage 4 (Self-Authoring Mind)} is organized through an internalized value system or self-authored framework that can evaluate and reject external rules. 
\textbf{Stage 5 (Self-Transforming Mind)} can reflect on and coordinate multiple systems at once, holding contradiction and paradox without collapsing them prematurely. 

In practice, respondents may also be rated as \emph{transitional} between adjacent stages (e.g., 3/4 or 4/5), indicating that their responses show meaningful evidence of both structures rather than a fully consolidated center of gravity at either one stage or the other.

Population estimates in Kegan’s framework place Stage 3 and Stage 4 as the most common adult structures, with Stage 2 and Stage 5 comparatively rare \cite{kegan1994}.


\subsection{Questionnaires and instrument comparison}
\label{app:questionnaires}

This section presents the DSCT and SCT questionnaires, along with the supplementary comparison showing that, on the simulated-persona benchmark, the shorter DSCT preserves enough signal for the strongest classifiers to recover stage as reliably as with the longer SCT.

\subsubsection{Developmental Sentence Completion Test (DSCT)}
\label{app:dsct-questionnaire}

The Developmental Sentence Completion Test (DSCT) is a 20-item self-administered instrument designed to elicit text samples sufficient for a trained human rater or LLM to make a tentative assessment of developmental structure. It is organized into two sections of 10 items each. The first section uses first-person prompts (\emph{self-assessment}); the second uses matched third-person prompts about a generic other (\emph{abstracted assessment}). Respondents were instructed to complete each sentence with the first thought that came to mind, using a single word, short phrase, or full paragraph.

\begin{nolinenumbers}
\begin{multicols}{2}
\small
\paragraph{Self-Assessment}
\begin{enumerate}
    \item When a promise is broken\ldots
    \item When both choices feel right\ldots
    \item When things don't go as I hoped\ldots
    \item When the hard work finally pays off\ldots
    \item If I am asked to compromise\ldots
    \item When I realized someone was paying attention\ldots
    \item Saying goodbye to something that mattered\ldots
    \item When what used to work no longer works\ldots
    \item When I have to choose what comes first\ldots
    \item When I realize I cannot control what happens next\ldots
\end{enumerate}
\small
\paragraph{Abstracted Assessment}
\begin{enumerate}
    \setcounter{enumi}{10}
    \item When a person feels they were treated unfairly by their supervisor\ldots
    \item If a person feels pulled between their own view and the expectations of others\ldots
    \item A person works very hard on something important, but it fails\ldots
    \item A team celebrates a project that succeeded. The person who led it\ldots
    \item A person believes a decision their group supports is wrong; they will\ldots
    \item When a person sees someone make a sacrifice for others\ldots
    \item When a person has to leave a role or place that was important to them\ldots
    \item A person realizes their plans need to change significantly. This might affect their existing beliefs\ldots
    \item When a person must choose between two opportunities that both seem meaningful\ldots
    \item A person must make an important decision and take on a new responsibility without complete information\ldots
\end{enumerate}
\end{multicols}
\end{nolinenumbers}
\subsubsection{Sentence Completion Test (SCT)}
\label{app:sct-questionnaire}

For comparison, we also include the longer 36-item Loevinger Sentence Completion Test (SCT), which served as the reference instrument in the DSCT vs.\ SCT comparison reported in Appendix~\ref{app:dsct-vs-sct}. As in DSCT, respondents were instructed to complete each stem with the first thought that came to mind.

\begin{nolinenumbers}
\begin{multicols}{2}
\small
\begin{enumerate}
    \item When a child will not join in group activities\ldots
    \item Raising a family\ldots
    \item When I am criticized\ldots
    \item A man's job\ldots
    \item Being with other people\ldots
    \item The thing I like about myself is\ldots
    \item My mother and I\ldots
    \item What gets me into trouble is\ldots
    \item Education\ldots
    \item When people are helpless\ldots
    \item Women are lucky because\ldots
    \item A good boss\ldots\ (Alternative: A good father\ldots)
    \item A girl/boy has a right to\ldots
    \item When they talked about sex, I\ldots
    \item A wife/husband should\ldots
    \item I feel sorry\ldots
    \item A man/woman feels good when\ldots
    \item Rules are\ldots
    \item Crime and delinquency could be halted if\ldots
    \item Men are lucky because\ldots
    \item I just can't stand people who\ldots
    \item At times she/he worried about\ldots
    \item I am\ldots
    \item A woman/man feels good when\ldots
    \item My main problem is\ldots
    \item Whenever she/he was with her/his mother, she/he\ldots
    \item The worst thing about being a woman/man is\ldots
    \item A good mother\ldots
    \item Sometimes she/he wished that\ldots
    \item When I am with a man/woman\ldots
    \item When she/he thought of her/his mother, she/he\ldots
    \item If I can't get what I want\ldots
    \item Usually she/he felt that sex\ldots
    \item For a woman/man, a career is\ldots
    \item My conscience bothers me if\ldots
    \item A woman/man should always\ldots
\end{enumerate}
\end{multicols}
\end{nolinenumbers}

\subsubsection{DSCT vs.\ SCT comparison}
\label{app:dsct-vs-sct}

DSCT was designed as a shorter, less invasive successor to the Loevinger SCT \citep{loevinger1998technical} (see Section~\ref{sec:dsct-design}). To check that DSCT preserves enough developmental signal for computational stage classification, we ran a parallel comparison on the simulated-persona set used in Experiment~1.

\textbf{Setup.} Each of the 23 developmental profiles produced six independent simulated response sets to the 36-item SCT and six independent simulated response sets to the 20-item DSCT, yielding 138 classification decisions per classifier per instrument. The simulator prompt and system instruction were identical across the two questionnaires; only the questionnaire stems differed. We then asked the two top-performing classifiers from Experiment~1, Gemini 3.1 Pro and Claude Opus 4.6, to classify each response set into a Kegan stage.

\textbf{Results.} Both classifiers correctly recovered the simulator-intended target stage on all 23 profiles, on all six rounds, for both questionnaires (Table~\ref{tab:dsct-sct-comparison}). In other words, on this simulated-persona benchmark, the shorter DSCT carried enough signal for the strongest classifiers to recover stage as reliably as the longer SCT.

\begin{table}[h]
\centering
\caption{Classifier accuracy against simulator-intended target stage, by questionnaire. Numerator is profiles correctly classified out of 23. Six rounds per profile; stage labels were identical across rounds in every cell.}
\label{tab:dsct-sct-comparison}
\begin{tabular}{lcc}
\toprule
Classifier & SCT (36 items) & DSCT (20 items) \\
\midrule
Gemini 3.1 Pro & 23 / 23 & 23 / 23 \\
Claude Opus 4.6 & 23 / 23 & 23 / 23 \\
\bottomrule
\end{tabular}
\end{table}

\textbf{Sub-stage nuance.} The simulator-intended labels for transitional profiles use the X/Y notation (e.g., 3/4), and our classifiers were instructed to output labels in that same form. Some source profiles in the developmental literature contain finer distinctions (e.g., 4(3) or 3(4)) \citep{laske2023process,baxtermagolda2007interview}. Our classifiers correctly recovered the broader transitional category, but not these asymmetric within-transition emphases. We treat this as expected given the coarser X/Y output format.

\textbf{Scope.} This comparison does not show that DSCT and SCT are interchangeable for human respondents, nor that DSCT covers the full construct space of SCT or SCTi-MAP \citep{cook1999postautonomous}. It supports the narrower claim that, on simulator-generated personas, the shorter DSCT preserves enough signal for the strongest available classifiers to recover the intended stage as reliably as with the longer SCT.

\subsection{Rater protocol}
\label{app:rater-protocol}

All human ratings in this paper were based on written responses to the DSCT and used Robert Kegan's constructive-developmental framework to infer meaning-making structure from text. Raters were instructed to focus on \emph{how} experience was organized in the response rather than on verbal sophistication, morality, intelligence, or general maturity. In particular, the rating guide emphasized the subject--object distinction: what the respondent appeared able to reflect on versus what still seemed to organize the response from within. The guide also included stage descriptions, transitional structures, and cautions about low-evidence cases such as brief, affect-only, or linguistically limited responses.

\textbf{Allowed labels.} Raters were allowed to assign any of the following labels: Stage~1, Stage~2, Stage~2/3, Stage~3, Stage~3/4, Stage~4, Stage~4/5, and Stage~5. Stage~1 was allowed in principle but was not expected in the datasets considered here.

\textbf{Experiment 1.} In the simulated-persona condition, two raters independently evaluated a stratified subset of 46 simulated cases, corresponding to two independently generated response sets for each of the 23 developmental profiles. Raters were blind to the simulator-intended target stage, the profile identity, and one another's ratings. The two ratings were never more than a half-stage apart. The final human reference label for each case was taken as the lower of the two ratings.

\textbf{Experiment 2.} In the human-participant condition, three raters independently evaluated all 83 DSCT response sets. Raters were blind to participant identity and to one another's ratings, but they knew they were evaluating human rather than simulated responses. When at least two raters agreed, that label was taken as the consensus. In three-way disagreement cases, the middle rating was used as the final reference label. This rule applied both when the ratings formed an adjacent progression (e.g., 2, 2/3, 3) and when they spanned a larger gap (e.g., 2, 2/3, 4), yielding a conservative center label.

\textbf{Confidence and low-evidence cases.} Raters also assigned a confidence level (Low / Medium / High). The guide instructed raters to lower confidence when responses were sparse, vague, brief, affect-only, inconsistent across prompts, or linguistically limited, including cases likely shaped by non-native English fluency. Mixed responses across prompts were treated as normal; raters were asked to infer the respondent's overall center of gravity rather than coding from isolated strong answers, and to use transitional labels when the evidence suggested a genuine mix of adjacent structures.

\subsection{Experiment 1 supplementary materials}
\subsubsection{Simulated developmental profiles}
\label{app:agent-profiles}

Experiment~1 used 23 literature-derived developmental profiles as conditioning templates for persona simulation. These profiles were not treated as ground-truth people, but as compact textual representations of stage-specific meaning-making structures drawn from prior developmental literature. Each profile was associated with a target stage label used for simulation and evaluation.

\begin{table}[h!]
\centering
\caption{Literature-derived developmental profiles used as conditioning templates in Experiment~1.}
\label{tab:simulated-profiles}
\small
\begin{tabularx}{\textwidth}{l c X X}
\toprule
\textbf{Profile ID} & \textbf{Target Stage} & \textbf{Title} & \textbf{Source} \\
\midrule
S2.1   & 2   & The Tactical Adherence to Rules & \citealp{bartone2002cognitive} \\
S3.1   & 3   & Walter the Teacher & \citealp{berger2024changing} \\
S3.2   & 3   & Gwen (College Student) & \citealp{baxtermagolda2007interview} \\
S3.3   & 3   & John (The Perfectionist Leader) & \citealp{berger2024changing} \\
S3.4   & 3   & The Cross-Pressured Leader & \citealp{bartone2002cognitive} \\
S4.1   & 4   & Mark (College Student) & \citealp{baxtermagolda2007interview} \\
S4.2   & 4   & Professional Executive Coach & \citealp{kegan1994} \\
S4.3   & 4   & Organizational Manager (Participant 4) & \citealp{berger2024changing} \\
S4.4   & 4   & Adult Educator & \citealp{kegan1994} \\
S4.5   & 4   & The Emergence of the Leader of Character & \citealp{bartone2002cognitive} \\
S5.1   & 5   & The Adaptive Leader (Conflict Resolution) & \citealp{berger2024changing} \\
S5.2   & 5   & The Interdependent Partner (Intimate Relationships) & \citealp{kegan1994} \\
S23.1  & 2/3 & The Shift to Institutional Loyalty & \citealp{bartone2002cognitive} \\
S34.1  & 3/4 & Manager B & \citealp{laske2023process} \\
S34.2  & 3/4 & College Students Navigating Beliefs & \citealp{baxtermagolda2007interview} \\
S34.3  & 3/4 & Instructional Coach & \citealp{laske2023process} \\
S34.4  & 3/4 & Manager B (Corporate Decision Maker) & \citealp{laske2023process} \\
S34.5  & 3/4 & Community Leader (Work Motivation) & \citealp{berger2024changing} \\
S34.6  & 3/4 & Client A (Autodidact Coaching) & \citealp{laske2023process} \\
S34.7  & 3/4 & Peter (The Torn Husband/Employee) & \citealp{kegan1994} \\
S45.1  & 4/5 & Higher Education Leader & \citealp{berger2024changing} \\
S45.2  & 4/5 & Helen (The Transitioning Executive) & \citealp{kegan1994} \\
S45.3  & 4/5 & The Evolving Academic (Intellectual Humility) & \citealp{kegan1994} \\
\bottomrule
\end{tabularx}
\end{table}

\subsubsection{Simulation and classification prompts}
\label{app:exp1-prompts}

This subsection reports the prompt templates used in Experiment~1. Full questionnaire contents are given in Appendix~\ref{app:dsct-questionnaire} and Appendix~\ref{app:sct-questionnaire}.

\paragraph{Persona-conditioned simulation prompt (DSCT response generation).}
The following user prompt template was used to generate simulated DSCT responses from literature-derived developmental profiles:

\begin{quote}
\small
``You are roleplaying as the following person. Adopt their persona, developmental model, worldview, iq, and tone of voice completely.

Profile:

\texttt{[PROFILE\_JSON]}

Please answer the following questionnaire from the perspective of this person.

Questionnaire:

\texttt{[DSCT\_QUESTIONNAIRE]}''
\end{quote}

This prompt was combined with the following system instruction:

\begin{quote}
\small
``You are an expert at persona adoption and psychological roleplay. You must answer the questions exactly as the described person would, reflecting their specific developmental stage, biases, knowledge, limitations and personality. Don't make it too long. Answer like the real person would. Make each response differentiated enough but use mostly plain language, don't make it excessively complex in neither writing or vocabulary.''
\end{quote}

\paragraph{DSCT classification prompt.}
The following prompt template was used to classify DSCT responses into Kegan stages:

\begin{quote}
\small
``You are an expert developmental psychologist specializing in Robert Kegan's Constructive-Developmental Theory.
Your task is to evaluate the cognitive development stage of several candidates based on their responses to a questionnaire.

The possible stages are 1, 2, 3, 4, 5, and transitional stages (e.g., 2/3, 3/4, 4/5).

Here is the questionnaire the candidates responded to:

\texttt{[DSCT\_QUESTIONNAIRE]}

Attached are the candidates' responses to evaluate.

Please evaluate each candidate and provide their ID, the assigned Kegan stage, and a detailed explanation of your reasoning in 1 line.

Here are some examples of responses we expect from your evaluation in json format:

\texttt{[EXAMPLE\_OUTPUT\_JSON]}''
\end{quote}

\paragraph{SCT classification prompt.}
For the DSCT vs.\ SCT comparison, the same classification template was used with the questionnaire field replaced by the SCT stems:

\begin{quote}
\small
``You are an expert developmental psychologist specializing in Robert Kegan's Constructive-Developmental Theory.
Your task is to evaluate the cognitive development stage of several candidates based on their responses to a questionnaire.

The possible stages are 1, 2, 3, 4, 5, and transitional stages (e.g., 2/3, 3/4, 4/5).

Here is the questionnaire the candidates responded to:

\texttt{[SCT\_QUESTIONNAIRE]}

Attached are the candidates' responses to evaluate.

Please evaluate each candidate and provide their ID, the assigned Kegan stage, and a detailed explanation of your reasoning in 1 line.

Here are some examples of responses we expect from your evaluation in json:

\texttt{[EXAMPLE\_OUTPUT\_JSON]}''
\end{quote}

\subsubsection{Detailed statistics}
\label{app:exp1-stats}

This subsection reports the additional statistical tests underlying the model-tier comparisons in Experiment~1.

\textbf{Stage type effect.} Aggregating classifiers by model tier (compact/fast: Claude 4.5 Haiku, GPT 5 Mini, Gemini 3.1 Flash; Frontier: all remaining models), a two-way ANOVA on classification accuracy with factors \emph{Model Tier} and \emph{Stage Type} (Solid vs.\ Transitional) showed a main effect of Stage Type. This effect remained significant both when Mistral 3 Large was excluded as an outlier ($p < 0.0001$) and when it was retained ($p < 0.001$).

\textbf{Compact/Fast degradation on transitional stages.} compact/fast models maintained 98.3\% accuracy on solid-stage profiles but dropped to 66.5\% on transitional-stage profiles. This difference was significant under Welch's t-test ($t = 5.12$, $p = 0.006$), indicating that the main loss of performance in smaller models was concentrated on transitional cases rather than solid stages.

\textbf{Directional bias by model tier.} Overestimation bias also differed by model tier. compact/fast models showed substantially larger upward bias ($M = 12.7\%$, $SD = 3.3$) than frontier models ($M = 1.8\%$, $SD = 2.9$). This difference was significant under Welch's t-test ($t = 6.07$, $p = 0.001$). Under these controlled synthetic conditions, smaller models therefore failed not only more often, but also in a more directionally consistent way.

\textbf{Outlier handling.} Mistral 3 Large was treated as an outlier in one version of the ANOVA because its performance was substantially lower than the rest of the model set, particularly on transitional stages. We therefore report the Stage Type effect both with and without Mistral included. The qualitative interpretation was unchanged in both cases.

\subsection{Experiment 2 supplementary materials}
\label{app:exp2-supp}

\subsubsection{Human study materials}
\label{app:user-study}

This subsection reproduces the participant-facing materials used for the human study in Experiment~2. Participants were recruited through personal, academic, and professional networks. Compensation was a \$15 or \texteuro15 Amazon gift card (or equivalent).

\paragraph{Enrollment / sign-up form.}
Participants first completed an enrollment form that collected their email address and consent to participate. The form explained that, after enrollment, an experimenter would assign each participant a Subject ID to keep subsequent responses anonymous. It also described the overall study as consisting of two short tasks: (1) a 20-item sentence-completion questionnaire and (2) a prompt-based task in which participants would ask their regular LLM chatbot to complete a cognitive assessment and report the numerical output. The form stated that the full study would take approximately 30 minutes.

The enrollment form also stated that participation was voluntary, that participants could stop or withdraw at any time without giving a reason, and that they could contact the experimenters if they wished to withdraw. It further noted that the study followed the Declaration of Helsinki ethics guidelines and allowed participants to indicate whether they wished to receive their individual results and the aggregated study results. Under the institution’s policies, this study did not require formal IRB review because it was classified as low-risk online survey research.

\paragraph{Questionnaire instructions shown before DSCT.}
Before completing the DSCT, participants saw the following instructions:

\begin{quote}
\small
``This questionnaire contains 20 sentence-completion prompts.

For the first 10 prompts, you will be asked to complete sentences about how you think or respond when certain situations happen to you.

For the next 10 prompts, you will be asked to complete sentences about how a person might think or respond in a similar situation.

Please finish each sentence with the first thought that comes to mind.

There are no right or wrong answers.

You may complete the sentences with a single word, a short phrase, or a full paragraph, whichever best expresses your thought.

Please complete every item. Do not skip any sentences.

Work at a steady pace, and try not to overthink your responses.''
\end{quote}

\paragraph{Anonymization and data handling.}
After enrollment, each participant was assigned a Subject ID by an experimenter. This ID was used to link responses across study components while keeping questionnaire responses separate from identifying contact information.

\paragraph{Additional study component.}
After completing the DSCT, participants were invited to complete a second prompt-based task involving their regular LLM chatbot, described in Appendix~\ref{app:chat-followup}.

\subsubsection{Additional exploratory analysis}
\label{app:exp2-random-forest}

To better understand why Experiment~2 was harder than the synthetic-persona setting of Experiment~1, we ran additional item-level analyses on both the simulated and human DSCT responses. Human raters had already noted that the synthetic responses appeared unusually consistent compared to the human data. To examine this more directly, we prompted Gemini 3.1 Pro to assign a Kegan stage to each questionnaire item independently, rather than rating each 20-item response set as a whole.

This analysis confirmed the hyper-consistency of the synthetic data. Simulated persona responses were highly uniform across items: responses generated from a Stage~4 persona, for example, were rated almost uniformly as Stage~4 at the item level. The same pattern did not hold for real human responses, where developmental signal was more unevenly distributed across questions.

\begin{figure}[h!]
    \centering
    \includegraphics[width=0.72\linewidth]{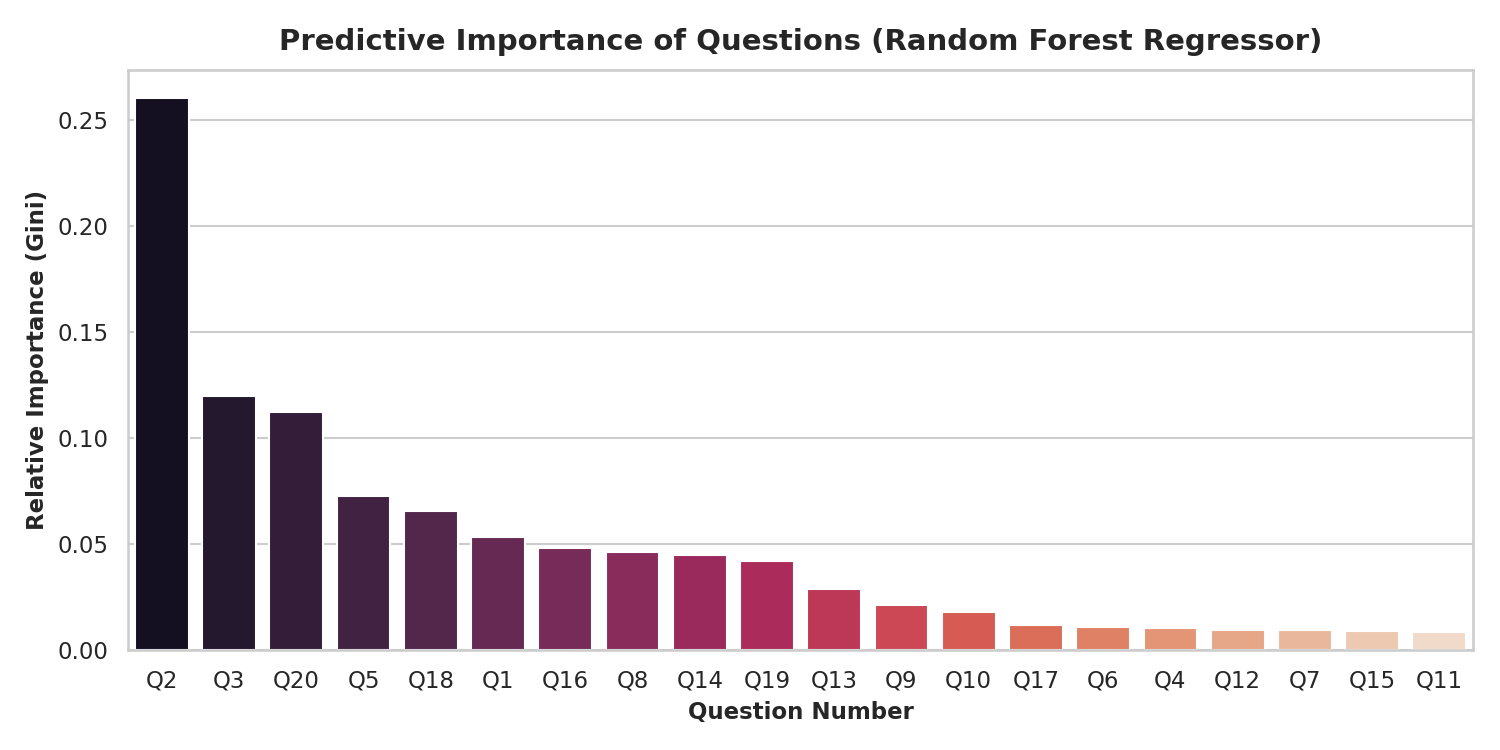}
    \caption{\textbf{Experiment 2 exploratory analysis.} Random forest feature importance across DSCT items for predicting final human stage labels. Predictive weight is concentrated in a small subset of questions, with Q2 contributing the largest share.}
    \label{fig:randomforest}
\end{figure}

In the human data, a Pearson correlation analysis identified Q2, Q20, Q3, Q16, and Q18 as the strongest item-level predictors of final stage labels ($r > 0.65$). By contrast, Q14, Q10, and Q8 appeared more susceptible to upward drift, with some Stage~3 respondents producing answers that resembled Stage~4-like logic on those prompts. The random forest analysis shown in Figure~\ref{fig:randomforest} supports the same interpretation: predictive importance was concentrated in a relatively small subset of items, with Q2 alone accounting for roughly 25\% of total feature importance.

These analyses suggest that real human DSCT responses contain usable developmental signal, but not with the near-uniform consistency seen in synthetic persona text. This helps explain why agreement in Experiment~2 is lower than in Experiment~1, and why the relevant benchmark target for human-written DSCT responses is approximate alignment with trained human judgment rather than exact stage recovery.

\subsection{Experiment 3 supplementary materials}
\subsubsection{Default DSCT prompt for models}
\label{app:exp3-prompts}

This subsection reports the prompt template used in Experiment~3, where models were asked to answer the DSCT without persona-conditioning or target-stage instructions. Full questionnaire contents are given in Appendix~\ref{app:dsct-questionnaire}.

\begin{quote}
\small
``This questionnaire contains 20 sentence-completion prompts.

For the first 10 prompts, you will be asked to complete sentences about how you think or respond when certain situations happen to you.

For the next 10 prompts, you will be asked to complete sentences about how a person might think or respond in a similar situation.

Please finish each sentence with the first thought that comes to mind.

There are no right or wrong answers.

You may complete the sentences with a single word, a short phrase, or a full paragraph, whichever best expresses your thought.

Please complete every item. Do not skip any sentences.

Work at a steady pace, and try not to overthink your responses.

Here is the questionnaire:

\texttt{[DSCT\_QUESTIONNAIRE]}

Here an example of the format for your responses in json:

\texttt{[RESPONSE\_FORMAT\_JSON]}''
\end{quote}

\subsubsection{Full ratings for model-generated DSCT responses}
\label{app:exp3-model-table}

Table~\ref{tab:exp3-model-stages} reports the full developmental ratings assigned to the DSCT responses generated by each model in Experiment~3. These ratings should be interpreted as descriptions of output structure under a shared elicitation format, not as claims about human-like developmental stages in the models themselves.

\begin{table}[htbp]
    \tiny
    \centering
    \caption{Developmental ratings assigned to default model-generated DSCT responses in Experiment~3.}
    \label{tab:exp3-model-stages}
    \begin{tabularx}{\textwidth}{p{2.2cm} p{0.8cm} X}
        \toprule
        \textbf{Model ID} & \textbf{Stage} & \textbf{Explanation} \\
        \midrule
        Claude Haiku 4.5 & 3/4 & The subject displays a transitional Stage 3/4 mindset, balancing a strong orientation toward mutual respect and harmony with emerging self-authoring traits like prioritizing long-term impacts and internal values. \\
        
        Claude Opus 4.6 & 5 & The subject exhibits a clear Stage 5 (Self-Transforming) consciousness, demonstrating dialectical thinking, the ability to objectify their own self-authoring systems, and a deep comfort with paradox, provisional framing, and systemic complexity. \\
        
        Claude Opus 4 & 4 & The subject demonstrates a solid Stage 4 (Self-Authoring) framework by consistently evaluating situations through an internal compass of core values, long-term goals, and systemic principles rather than merely reacting to external expectations. \\
        
        DeepSeek R1 & 4 & The subject indicates a Stage 4 meaning-making system through concise responses that prioritize internal principles, personal purpose, and the capacity to objectify external expectations to stay true to themselves. \\
        
        DeepSeek V4 & 4 & The subject reflects a strong Stage 4 development, consistently referencing core principles, self-authored values, and the ability to view social harmony as an object to be weighed against personal authenticity. \\
        
        Gemini 3.1 Flash & 4 & The subject operates primarily at Stage 4, evaluating ethical compromises and social pressures against a self-authored value system, while showing glimpses of transitioning to Stage 5 by acknowledging the need to adapt their underlying worldview. \\
        
        Gemini 3.1 Pro & 4 & The subject displays a Stage 4 (Self-Authoring) level of development, consistently relying on a self-generated sense of purpose and internal values to navigate interpersonal conflicts and cognitive adjustments. \\
        
        Gemini 2.5 Flash & 4 & The subject's highly analytical and systems-oriented responses indicate a Stage 4 mindset that relies on internal operational protocols, defined primary goals, and the ability to re-evaluate one's own identity as an object. \\
        
        Gemini 2.0 & 3/4 & The subject exhibits a transitional Stage 3/4 profile, showing the analytical capacity to weigh opposing perspectives while still developing the strong, independent value system characteristic of full self-authorship. \\
        
        Gemma 4 & 4 & The subject clearly operates at Stage 4, guided by core objectives and self-authored authenticity, while possessing the capacity to objectify and update their own perspectives when faced with new realities. \\
        
        Gemma 3 & 3 & The subject reflects a classic Stage 3 (Socialized Mind) perspective, being highly embedded in interpersonal expectations, paralyzed by a search for external rightness, and ultimately deferring to group cohesion. \\
        
        GPT-5.5 & 4/5 & The subject demonstrates a late Stage 4 to transitioning Stage 4/5 maturity, firmly rooted in self-authored principles while displaying a fluid, evolutionary understanding of their own unfolding identity and assumptions. \\
        
        GPT-5.4 & 4 & Exhibits robust self-authorship by establishing clear boundaries around `non-negotiable' values and possessing the systemic capacity to recognize when it has `outgrown an old pattern'. \\
        
        GPT-5 Mini & 3/4 & The subject reflects a Stage 3/4 transition, moving away from pure socialized expectations by relying on internal intuition and attempting to balance external demands with emerging personal authenticity. \\
        
        GPT-4.1 Mini & 3/4 & The subject is characteristic of a Stage 3/4 transition, demonstrating an emerging self-authored intuition while remaining significantly invested in balancing personal views with the expectations of the group. \\
        
        GPT-3.5 Turbo & 2 & Operates at the Imperial Mind stage; responses are entirely reduced to single-word, emotionally reactive, or purely transactional outcomes (`Disappointed', `Relieved', `Negotiate', `Oppose') without articulating a broader social or self-authored framework. \\
        
        Grok 4.2 & 4 & The subject exhibits a solid Stage 4 framework by relying on objective logic, self-determined priorities, and the capacity to systematically adapt their own beliefs to align with changing realities. \\
        
        Grok 3 & 4 & The subject operates at a Stage 4 level, utilizing self-authored long-term goals and personal values as the ultimate arbiters when navigating compromises and interpersonal perspectives. \\
        
        Grok 3 Mini & 3/4 & The subject displays a Stage 3/4 transitional orientation, showing a growing reliance on personal principles while still experiencing significant internal conflict when those principles clash with external expectations. \\
        
        Qwen 3.6 Plus & 4 & The subject represents a clear Stage 4 consciousness, consistently making sense of challenges through an internal compass of deeper priorities and self-authored alignment rather than external validation. \\
        
        Qwen 3 Max & 4 & The subject firmly demonstrates a Stage 4 (Self-Authoring) system, evaluating choices through core principles and capable of objectifying their own assumptions to integrate new perspectives into an internally coherent worldview. \\
        
        Qwen 2.5 Max & 3/4 & The subject reveals a Stage 3/4 profile, evidencing strong self-authored priorities while still exhibiting significant vulnerability to socialized fears, such as complying merely to avoid interpersonal isolation. \\
        
        Kimi K2.6 & 4 & The subject operates robustly at Stage 4, characterized by an ability to treat their own worldview as a cognitive model to be updated and prioritizing internal integrity over socialized loyalty. \\
        
        Meta AI Muse Spark & 4/5 & The subject demonstrates a highly advanced Stage 4/5 perspective, treating identity and beliefs as malleable constructs while analyzing systemic, second-order effects rather than merely reacting to them. \\
        
        Llama 3 & 3 & Responses are brief, externally oriented, and reactive (`I feel disappointed', `They may feel undervalued'), showing a Socialized Mind deeply embedded in immediate emotional and interpersonal consequences without a clear self-authored framework. \\
        
        Llama 2 & 3 & Exhibits a Socialized Mind; responses reflect an internalized drive for relational harmony (`find a solution that meets everyone's needs', `persuade the group to see things from their perspective') and rely on external `outcomes' rather than an independent value structure. \\
        
        Perplexity Sonar & 4 & The subject reflects a Stage 4 (Self-Authoring) level, anchoring their decisions in personal values and long-term meaning while acknowledging the internal tension involved in asserting this self-authored perspective. \\
        
        Mistral Large 3 & 4 & The subject consistently demonstrates a Stage 4 mindset, prioritizing their own convictions and self-authored values over group expectations, even while remaining open to questioning those beliefs. \\
        
        Mistral Medium 3.1 & 4/5 & The subject exhibits a late Stage 4 or transitional Stage 4/5 consciousness, possessing a firmly self-authored value system while demonstrating the self-transforming ability to objectify and debug their own beliefs as fluid, evolving constructs. \\
        \bottomrule
    \end{tabularx}
\end{table}

\subsection{Follow-up: stage inference from naturalistic conversation}
\label{app:chat-followup}

As a follow-up analysis, we compared stage assignments derived from participants' naturalistic chat history with stage assignments derived from the structured DSCT. This was not part of the core benchmark, but serves as a stress test of whether the same developmental signal remains recoverable when the input shifts from elicited questionnaire responses to unconstrained conversational traces.

We focus on the matched subset for which all three quantities were available ($N=69$): LLM grading of chat history, human grading of the questionnaire, and LLM grading of the questionnaire. Mean assigned stage differed substantially across these three conditions: LLM grading of chat history produced the highest mean stage ($M=4.15$, $SD=0.40$), followed by human grading of the questionnaire ($M=3.67$, $SD=0.51$), and LLM grading of the questionnaire ($M=3.53$, $SD=0.53$). A Friedman test showed a significant omnibus difference across the three paired conditions ($\chi^2(2)=52.435$, $p<.001$).

Agreement analyses revealed a pronounced conversational overestimation effect. Comparing LLM grading of chat history to human grading of the questionnaire yielded very weak agreement ($\kappa=0.022$), with 26.1\% exact agreement and 40.6\% severe disagreements (defined as $\geq 1.0$ full stage apart). Comparing LLM grading of chat history to LLM grading of the questionnaire also showed poor agreement ($\kappa=0.149$), with 18.8\% exact agreement and 39.1\% severe disagreements. By contrast, agreement between human and LLM grading of the questionnaire was substantially stronger ($\kappa=0.324$), with 43.5\% exact agreement and 73.9\% agreement within $\pm 0.5$ stage.

Refusal rates were also uneven across providers in the broader chat-history collection: 11 of 77 responses (14.3\%) did not produce a stage assignment, including 9 of 39 ChatGPT responses. This suggests that willingness to infer stage from prior chat history did not consistently track the actual availability of usable evidence.

These results suggest that the strongest source of upward bias is not the classifier alone, but the conversational medium itself: the same participant can appear substantially more ``self-authored'' when evaluated through unstructured chat history than when evaluated through structured DSCT responses.

\begin{figure}[h!]
    \centering
    \includegraphics[width=0.72\linewidth]{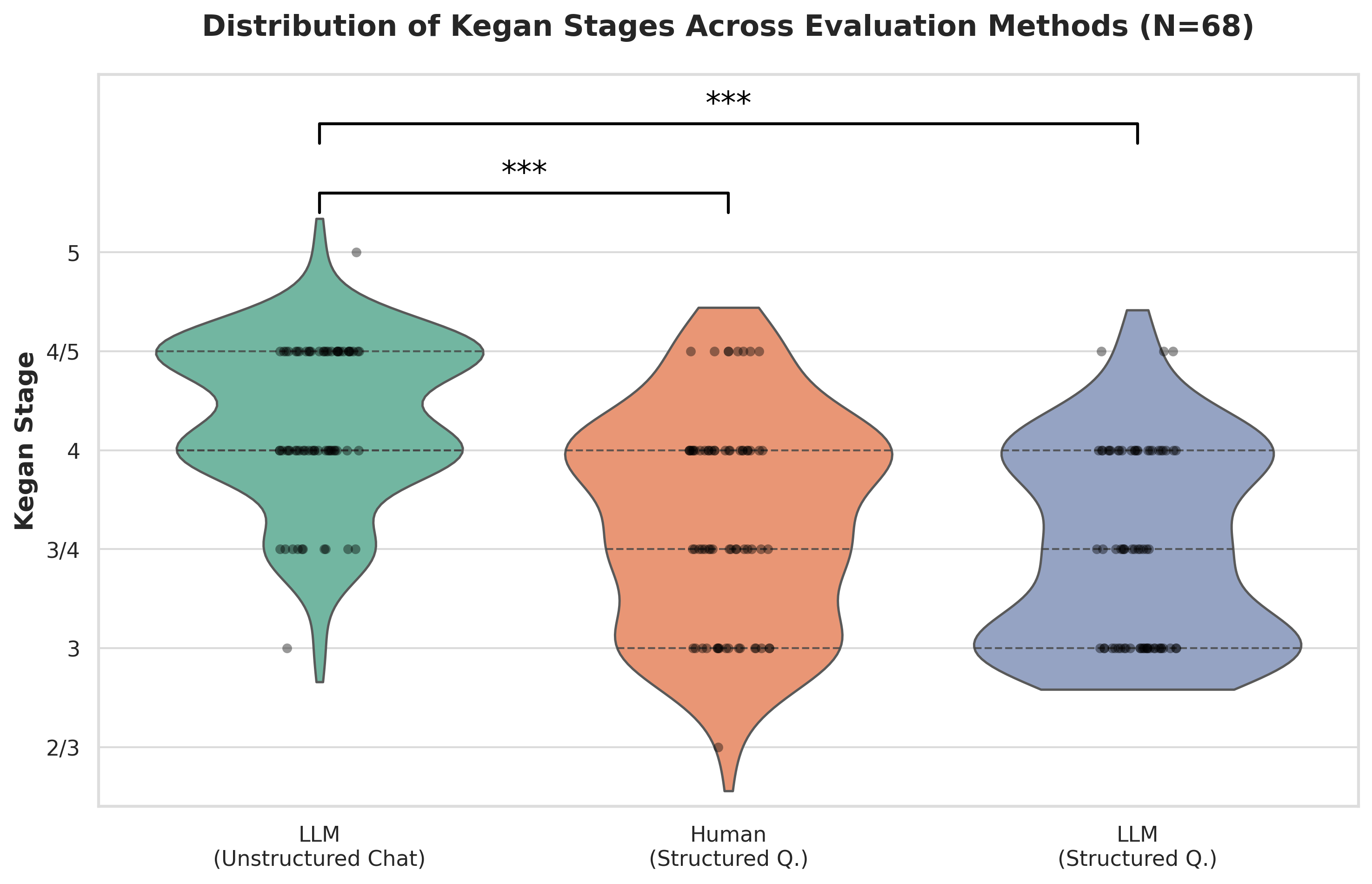}
    \caption{\textbf{Stage distributions across the matched $N=69$ subset.} LLM grading of chat history is systematically shifted upward relative to both human and LLM grading of the structured questionnaire.}
    \label{fig:chat_followup_violin}
\end{figure}

\begin{figure}[h!]

    \begin{minipage}[t]{0.5\linewidth}
        \centering
        \includegraphics[width=\linewidth]{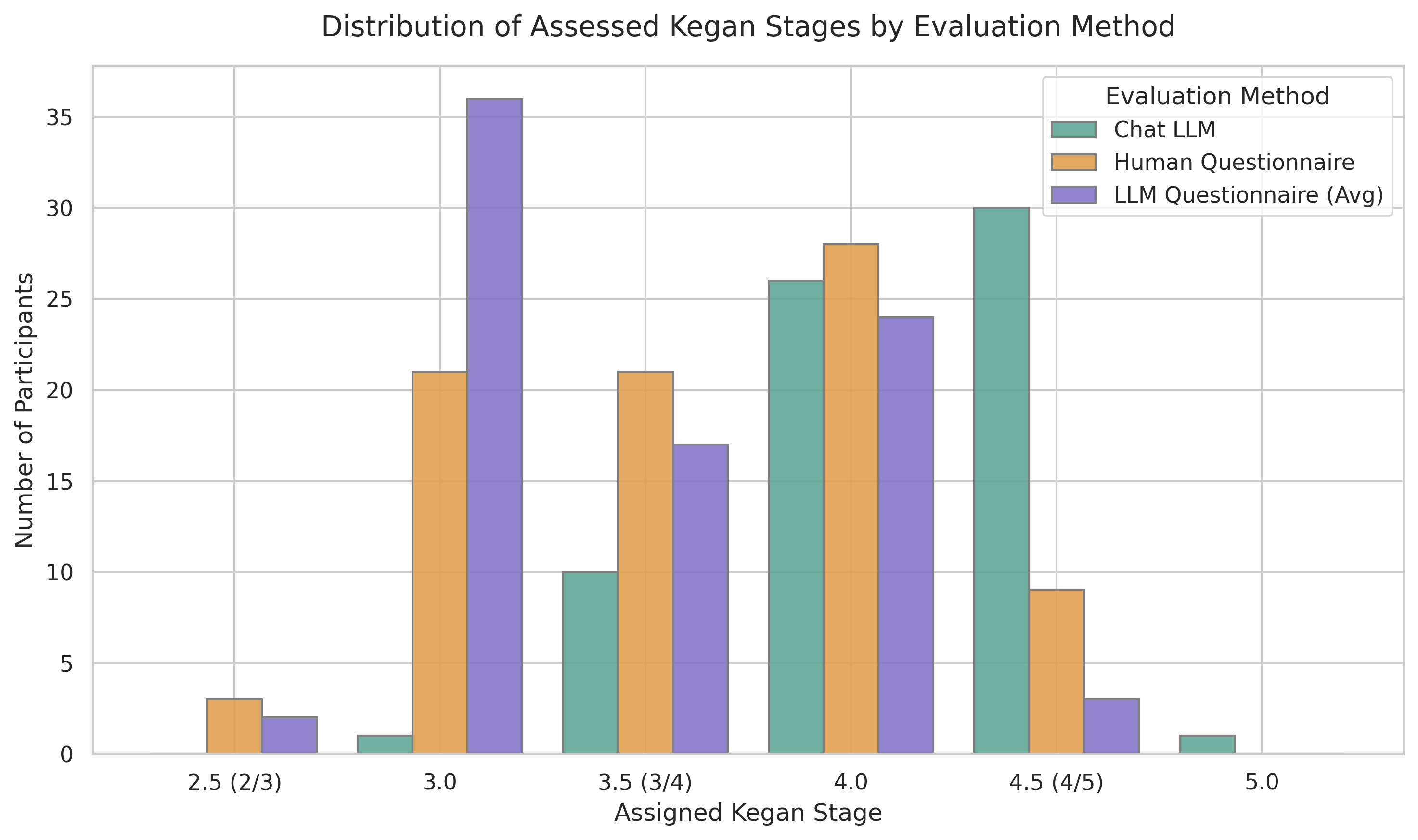}
    \end{minipage}\hfill
\centering
    \begin{minipage}[t]{0.35\linewidth}
    
    \includegraphics[width=\linewidth]{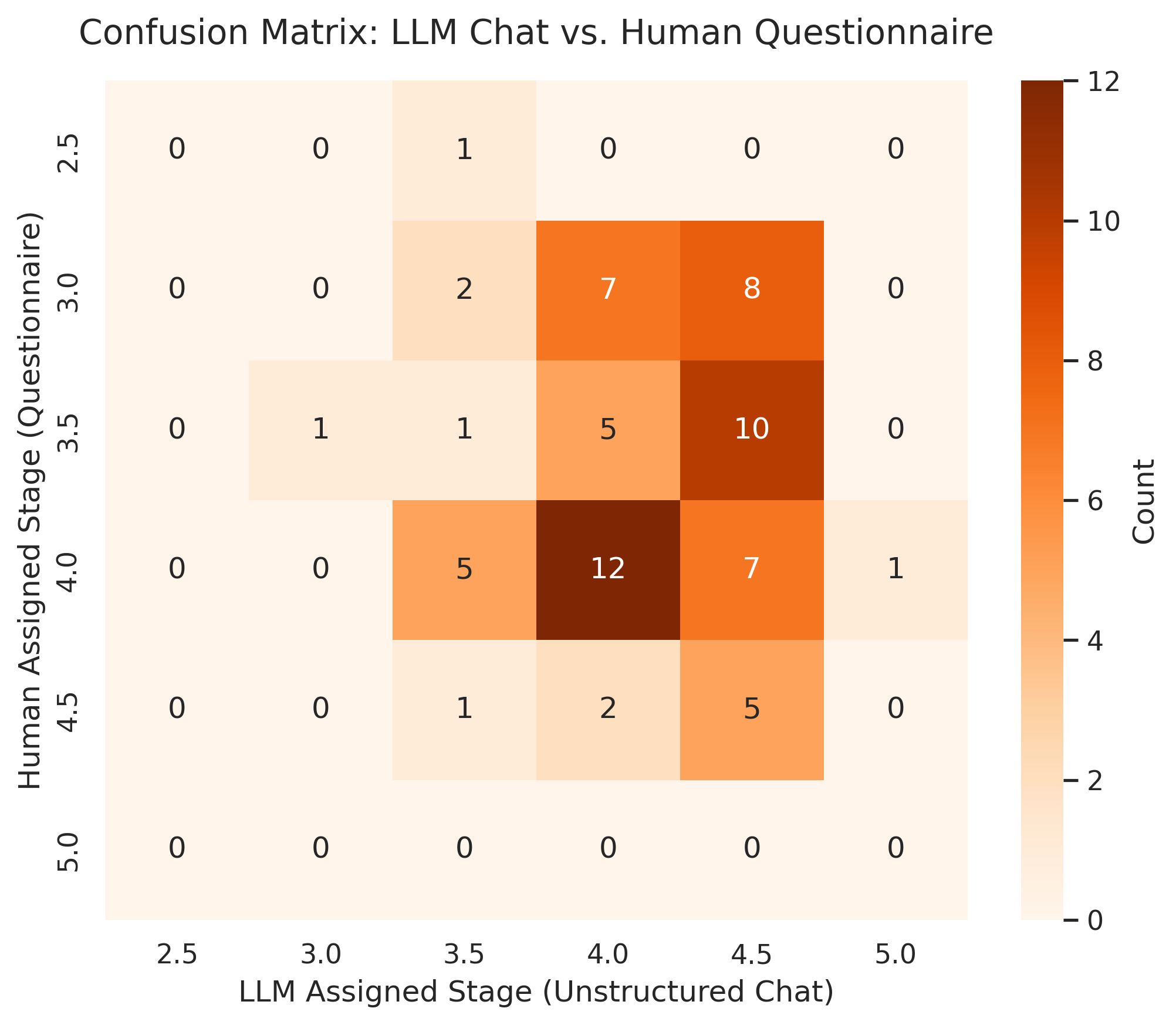}
     \end{minipage}
    \label{fig:chat_followup_heatmaps}
     \caption{\textbf{Scoring and Agreement heatmaps for conversational and questionnaire-based stage inference.} LLMs using conversational data over estimate the kegan stage of their users. Agreement is very weak for LLM(chat) vs.\ Human(questionnaire).}
\end{figure}


\newpage
\section*{NeurIPS Paper Checklist}

The checklist is designed to encourage best practices for responsible machine learning research, addressing issues of reproducibility, transparency, research ethics, and societal impact. Do not remove the checklist: {\bf The papers not including the checklist will be desk rejected.} The checklist should follow the references and follow the (optional) supplemental material.  The checklist does NOT count towards the page
limit. 

Please read the checklist guidelines carefully for information on how to answer these questions. For each question in the checklist:
\begin{itemize}
    \item You should answer \answerYes{}, \answerNo{}, or \answerNA{}.
    \item \answerNA{} means either that the question is Not Applicable for that particular paper or the relevant information is Not Available.
    \item Please provide a short (1--2 sentence) justification right after your answer (even for \answerNA). 
\end{itemize}

{\bf The checklist answers are an integral part of your paper submission.} They are visible to the reviewers, area chairs, senior area chairs, and ethics reviewers. You will also be asked to include it (after eventual revisions) with the final version of your paper, and its final version will be published with the paper.

The reviewers of your paper will be asked to use the checklist as one of the factors in their evaluation. While \answerYes{} is generally preferable to \answerNo{}, it is perfectly acceptable to answer \answerNo{} provided a proper justification is given (e.g., error bars are not reported because it would be too computationally expensive'' or ``we were unable to find the license for the dataset we used''). In general, answering \answerNo{} or \answerNA{} is not grounds for rejection. While the questions are phrased in a binary way, we acknowledge that the true answer is often more nuanced, so please just use your best judgment and write a justification to elaborate. All supporting evidence can appear either in the main paper or the supplemental material, provided in appendix. If you answer \answerYes{} to a question, in the justification please point to the section(s) where related material for the question can be found.

IMPORTANT, please:
\begin{itemize}
    \item {\bf Delete this instruction block, but keep the section heading ``NeurIPS Paper Checklist"},
    \item  {\bf Keep the checklist subsection headings, questions/answers and guidelines below.}
    \item {\bf Do not modify the questions and only use the provided macros for your answers}.
\end{itemize}


\begin{enumerate}

\item {\bf Claims}
    \item[] Question: Do the main claims made in the abstract and introduction accurately reflect the paper's contributions and scope?
    \item[] Answer: \answerYes{}
    \item[] Justification: All the claims are justified via experimentation with frontier LLMs and Human studies. 
    \item[] Guidelines:
    \begin{itemize}
        \item The answer \answerNA{} means that the abstract and introduction do not include the claims made in the paper.
        \item The abstract and/or introduction should clearly state the claims made, including the contributions made in the paper and important assumptions and limitations. A \answerNo{} or \answerNA{} answer to this question will not be perceived well by the reviewers. 
        \item The claims made should match theoretical and experimental results, and reflect how much the results can be expected to generalize to other settings. 
        \item It is fine to include aspirational goals as motivation as long as it is clear that these goals are not attained by the paper. 
    \end{itemize}

\item {\bf Limitations}
    \item[] Question: Does the paper discuss the limitations of the work performed by the authors?
    \item[] Answer: \answerYes{}
    \item[] Justification:  We include a through limitation discussion.
    \item[] Guidelines:
    \begin{itemize}
        \item The answer \answerNA{} means that the paper has no limitation while the answer \answerNo{} means that the paper has limitations, but those are not discussed in the paper. 
        \item The authors are encouraged to create a separate ``Limitations'' section in their paper.
        \item The paper should point out any strong assumptions and how robust the results are to violations of these assumptions (e.g., independence assumptions, noiseless settings, model well-specification, asymptotic approximations only holding locally). The authors should reflect on how these assumptions might be violated in practice and what the implications would be.
        \item The authors should reflect on the scope of the claims made, e.g., if the approach was only tested on a few datasets or with a few runs. In general, empirical results often depend on implicit assumptions, which should be articulated.
        \item The authors should reflect on the factors that influence the performance of the approach. For example, a facial recognition algorithm may perform poorly when image resolution is low or images are taken in low lighting. Or a speech-to-text system might not be used reliably to provide closed captions for online lectures because it fails to handle technical jargon.
        \item The authors should discuss the computational efficiency of the proposed algorithms and how they scale with dataset size.
        \item If applicable, the authors should discuss possible limitations of their approach to address problems of privacy and fairness.
        \item While the authors might fear that complete honesty about limitations might be used by reviewers as grounds for rejection, a worse outcome might be that reviewers discover limitations that aren't acknowledged in the paper. The authors should use their best judgment and recognize that individual actions in favor of transparency play an important role in developing norms that preserve the integrity of the community. Reviewers will be specifically instructed to not penalize honesty concerning limitations.
    \end{itemize}

\item {\bf Theory assumptions and proofs}
    \item[] Question: For each theoretical result, does the paper provide the full set of assumptions and a complete (and correct) proof?
    \item[] Answer:  \answerNA{}.
    \item[] Justification: There’s no theoretical results in this paper.
    \item[] Guidelines:
    \begin{itemize}
        \item The answer \answerNA{} means that the paper does not include theoretical results. 
        \item All the theorems, formulas, and proofs in the paper should be numbered and cross-referenced.
        \item All assumptions should be clearly stated or referenced in the statement of any theorems.
        \item The proofs can either appear in the main paper or the supplemental material, but if they appear in the supplemental material, the authors are encouraged to provide a short proof sketch to provide intuition. 
        \item Inversely, any informal proof provided in the core of the paper should be complemented by formal proofs provided in appendix or supplemental material.
        \item Theorems and Lemmas that the proof relies upon should be properly referenced. 
    \end{itemize}

    \item {\bf Experimental result reproducibility}
    \item[] Question: Does the paper fully disclose all the information needed to reproduce the main experimental results of the paper to the extent that it affects the main claims and/or conclusions of the paper (regardless of whether the code and data are provided or not)?
    \item[] Answer: \answerTODO{} \answerYes{}
    \item[] Justification: The paper fully disclose all the information needed to reproduce the main
experimental results of the paper, and we release all our prompts, datasets and evaluation tools to assist the reproducibility
of our experimental results.
    \item[] Guidelines:
    \begin{itemize}
        \item The answer \answerNA{} means that the paper does not include experiments.
        \item If the paper includes experiments, a \answerNo{} answer to this question will not be perceived well by the reviewers: Making the paper reproducible is important, regardless of whether the code and data are provided or not.
        \item If the contribution is a dataset and\slash or model, the authors should describe the steps taken to make their results reproducible or verifiable. 
        \item Depending on the contribution, reproducibility can be accomplished in various ways. For example, if the contribution is a novel architecture, describing the architecture fully might suffice, or if the contribution is a specific model and empirical evaluation, it may be necessary to either make it possible for others to replicate the model with the same dataset, or provide access to the model. In general. releasing code and data is often one good way to accomplish this, but reproducibility can also be provided via detailed instructions for how to replicate the results, access to a hosted model (e.g., in the case of a large language model), releasing of a model checkpoint, or other means that are appropriate to the research performed.
        \item While NeurIPS does not require releasing code, the conference does require all submissions to provide some reasonable avenue for reproducibility, which may depend on the nature of the contribution. For example
        \begin{enumerate}
            \item If the contribution is primarily a new algorithm, the paper should make it clear how to reproduce that algorithm.
            \item If the contribution is primarily a new model architecture, the paper should describe the architecture clearly and fully.
            \item If the contribution is a new model (e.g., a large language model), then there should either be a way to access this model for reproducing the results or a way to reproduce the model (e.g., with an open-source dataset or instructions for how to construct the dataset).
            \item We recognize that reproducibility may be tricky in some cases, in which case authors are welcome to describe the particular way they provide for reproducibility. In the case of closed-source models, it may be that access to the model is limited in some way (e.g., to registered users), but it should be possible for other researchers to have some path to reproducing or verifying the results.
        \end{enumerate}
    \end{itemize}

\item {\bf Open access to data and code}
    \item[] Question: Does the paper provide open access to the data and code, with sufficient instructions to faithfully reproduce the main experimental results, as described in supplemental material?
    \item[] Answer: \answerTODO{}\answerYes{}
    \item[] Justification:  This paper provides open access to the data, evaluation prompts and results and include instructions
for running the experiments. We include the links to the dataset collection.

    \item[] Guidelines:
    \begin{itemize}
        \item The answer \answerNA{} means that paper does not include experiments requiring code.
        \item Please see the NeurIPS code and data submission guidelines (\url{https://neurips.cc/public/guides/CodeSubmissionPolicy}) for more details.
        \item While we encourage the release of code and data, we understand that this might not be possible, so \answerNo{} is an acceptable answer. Papers cannot be rejected simply for not including code, unless this is central to the contribution (e.g., for a new open-source benchmark).
        \item The instructions should contain the exact command and environment needed to run to reproduce the results. See the NeurIPS code and data submission guidelines (\url{https://neurips.cc/public/guides/CodeSubmissionPolicy}) for more details.
        \item The authors should provide instructions on data access and preparation, including how to access the raw data, preprocessed data, intermediate data, and generated data, etc.
        \item The authors should provide scripts to reproduce all experimental results for the new proposed method and baselines. If only a subset of experiments are reproducible, they should state which ones are omitted from the script and why.
        \item At submission time, to preserve anonymity, the authors should release anonymized versions (if applicable).
        \item Providing as much information as possible in supplemental material (appended to the paper) is recommended, but including URLs to data and code is permitted.
    \end{itemize}

\item {\bf Experimental setting/details}
    \item[] Question: Does the paper specify all the training and test details (e.g., data splits, hyperparameters, how they were chosen, type of optimizer) necessary to understand the results?
    \item[] Answer: \answerNA{}.
    \item[] Justification: No training was performed.
    \item[] Guidelines:
    \begin{itemize}
        \item The answer \answerNA{} means that the paper does not include experiments.
        \item The experimental setting should be presented in the core of the paper to a level of detail that is necessary to appreciate the results and make sense of them.
        \item The full details can be provided either with the code, in appendix, or as supplemental material.
    \end{itemize}

\item {\bf Experiment statistical significance}
    \item[] Question: Does the paper report error bars suitably and correctly defined or other appropriate information about the statistical significance of the experiments?
    \item[] Answer: \answerYes{}
    \item[] Justification: We provide statistical significance analyses of the Pearson correlation, Wilcoxon rank tests, Cohen's Kappas and in general differences between the full sets of data for all experiments.

    \item[] Guidelines:
    \begin{itemize}
        \item The answer \answerNA{} means that the paper does not include experiments.
        \item The authors should answer \answerYes{} if the results are accompanied by error bars, confidence intervals, or statistical significance tests, at least for the experiments that support the main claims of the paper.
        \item The factors of variability that the error bars are capturing should be clearly stated (for example, train/test split, initialization, random drawing of some parameter, or overall run with given experimental conditions).
        \item The method for calculating the error bars should be explained (closed form formula, call to a library function, bootstrap, etc.)
        \item The assumptions made should be given (e.g., Normally distributed errors).
        \item It should be clear whether the error bar is the standard deviation or the standard error of the mean.
        \item It is OK to report 1-sigma error bars, but one should state it. The authors should preferably report a 2-sigma error bar than state that they have a 96\% CI, if the hypothesis of Normality of errors is not verified.
        \item For asymmetric distributions, the authors should be careful not to show in tables or figures symmetric error bars that would yield results that are out of range (e.g., negative error rates).
        \item If error bars are reported in tables or plots, the authors should explain in the text how they were calculated and reference the corresponding figures or tables in the text.
    \end{itemize}

\item {\bf Experiments compute resources}
    \item[] Question: For each experiment, does the paper provide sufficient information on the computer resources (type of compute workers, memory, time of execution) needed to reproduce the experiments?
    \item[] Answer: \answerTODO{} \answerYes{}
    \item[] Justification:  We include all experiments and human annotation resources in the Appendix, covering all resources used for generating model responses and evaluations,
and for collecting human labels.
    \item[] Guidelines:
    \begin{itemize}
        \item The answer \answerNA{} means that the paper does not include experiments.
        \item The paper should indicate the type of compute workers CPU or GPU, internal cluster, or cloud provider, including relevant memory and storage.
        \item The paper should provide the amount of compute required for each of the individual experimental runs as well as estimate the total compute. 
        \item The paper should disclose whether the full research project required more compute than the experiments reported in the paper (e.g., preliminary or failed experiments that didn't make it into the paper). 
    \end{itemize}
    
\item {\bf Code of ethics}
    \item[] Question: Does the research conducted in the paper conform, in every respect, with the NeurIPS Code of Ethics \url{https://neurips.cc/public/EthicsGuidelines}?
    \item[] Answer: \answerYes{}
    \item[] Justification: We confirm the research conducted in the paper conform, in every respect,
with the NeurIPS Code of Ethics.
    \item[] Guidelines:
    \begin{itemize}
        \item The answer \answerNA{} means that the authors have not reviewed the NeurIPS Code of Ethics.
        \item If the authors answer \answerNo, they should explain the special circumstances that require a deviation from the Code of Ethics.
        \item The authors should make sure to preserve anonymity (e.g., if there is a special consideration due to laws or regulations in their jurisdiction).
    \end{itemize}

\item {\bf Broader impacts}
    \item[] Question: Does the paper discuss both potential positive societal impacts and negative societal impacts of the work performed?
    \item[] Answer: \answerYes{}
    \item[] Justification: We discuss impacts and limitations both positive and negative in the discussion.
    \item[] Guidelines:
    \begin{itemize}
        \item The answer \answerNA{} means that there is no societal impact of the work performed.
        \item If the authors answer \answerNA{} or \answerNo, they should explain why their work has no societal impact or why the paper does not address societal impact.
        \item Examples of negative societal impacts include potential malicious or unintended uses (e.g., disinformation, generating fake profiles, surveillance), fairness considerations (e.g., deployment of technologies that could make decisions that unfairly impact specific groups), privacy considerations, and security considerations.
        \item The conference expects that many papers will be foundational research and not tied to particular applications, let alone deployments. However, if there is a direct path to any negative applications, the authors should point it out. For example, it is legitimate to point out that an improvement in the quality of generative models could be used to generate Deepfakes for disinformation. On the other hand, it is not needed to point out that a generic algorithm for optimizing neural networks could enable people to train models that generate Deepfakes faster.
        \item The authors should consider possible harms that could arise when the technology is being used as intended and functioning correctly, harms that could arise when the technology is being used as intended but gives incorrect results, and harms following from (intentional or unintentional) misuse of the technology.
        \item If there are negative societal impacts, the authors could also discuss possible mitigation strategies (e.g., gated release of models, providing defenses in addition to attacks, mechanisms for monitoring misuse, mechanisms to monitor how a system learns from feedback over time, improving the efficiency and accessibility of ML).
    \end{itemize}
    
\item {\bf Safeguards}
    \item[] Question: Does the paper describe safeguards that have been put in place for responsible release of data or models that have a high risk for misuse (e.g., pre-trained language models, image generators, or scraped datasets)?
    \item[] Answer:  \answerYes{}
    \item[] Justification:  We discuss safeguards  for responsible release use of the findings and questionnaires for evaluating developmental cognition.
    \item[] Guidelines:
    \begin{itemize}
        \item The answer \answerNA{} means that the paper poses no such risks.
        \item Released models that have a high risk for misuse or dual-use should be released with necessary safeguards to allow for controlled use of the model, for example by requiring that users adhere to usage guidelines or restrictions to access the model or implementing safety filters. 
        \item Datasets that have been scraped from the Internet could pose safety risks. The authors should describe how they avoided releasing unsafe images.
        \item We recognize that providing effective safeguards is challenging, and many papers do not require this, but we encourage authors to take this into account and make a best faith effort.
    \end{itemize}

\item {\bf Licenses for existing assets}
    \item[] Question: Are the creators or original owners of assets (e.g., code, data, models), used in the paper, properly credited and are the license and terms of use explicitly mentioned and properly respected?
    \item[] Answer: \answerYes{}
    \item[] Justification: The creators or original owners of assets used in the paper are properly credited
and are respected for the license and terms of use explicitly mentioned.
    \item[] Guidelines:
    \begin{itemize}
        \item The answer \answerNA{} means that the paper does not use existing assets.
        \item The authors should cite the original paper that produced the code package or dataset.
        \item The authors should state which version of the asset is used and, if possible, include a URL.
        \item The name of the license (e.g., CC-BY 4.0) should be included for each asset.
        \item For scraped data from a particular source (e.g., website), the copyright and terms of service of that source should be provided.
        \item If assets are released, the license, copyright information, and terms of use in the package should be provided. For popular datasets, \url{paperswithcode.com/datasets} has curated licenses for some datasets. Their licensing guide can help determine the license of a dataset.
        \item For existing datasets that are re-packaged, both the original license and the license of the derived asset (if it has changed) should be provided.
        \item If this information is not available online, the authors are encouraged to reach out to the asset's creators.
    \end{itemize}

\item {\bf New assets}
    \item[] Question: Are new assets introduced in the paper well documented and is the documentation provided alongside the assets?
    \item[] Answer: \answerYes{}
    \item[] Justification: We document all assets.
    \item[] Guidelines:
    \begin{itemize}
        \item The answer \answerNA{} means that the paper does not release new assets.
        \item Researchers should communicate the details of the dataset\slash code\slash model as part of their submissions via structured templates. This includes details about training, license, limitations, etc. 
        \item The paper should discuss whether and how consent was obtained from people whose asset is used.
        \item At submission time, remember to anonymize your assets (if applicable). You can either create an anonymized URL or include an anonymized zip file.
    \end{itemize}

\item {\bf Crowdsourcing and research with human subjects}
    \item[] Question: For crowdsourcing experiments and research with human subjects, does the paper include the full text of instructions given to participants and screenshots, if applicable, as well as details about compensation (if any)? 
    \item[] Answer: \answerYes{}
    \item[] Justification:  We include details for human annotations and questionnaires used in the Appendix.
    \item[] Guidelines:
    \begin{itemize}
        \item The answer \answerNA{} means that the paper does not involve crowdsourcing nor research with human subjects.
        \item Including this information in the supplemental material is fine, but if the main contribution of the paper involves human subjects, then as much detail as possible should be included in the main paper. 
        \item According to the NeurIPS Code of Ethics, workers involved in data collection, curation, or other labor should be paid at least the minimum wage in the country of the data collector. 
    \end{itemize}

\item {\bf Institutional review board (IRB) approvals or equivalent for research with human subjects}
    \item[] Question: Does the paper describe potential risks incurred by study participants, whether such risks were disclosed to the subjects, and whether Institutional Review Board (IRB) approvals (or an equivalent approval/review based on the requirements of your country or institution) were obtained?
    \item[] Answer: \answerNA{}.
    \item[] Justification:  Our human annotation is innocuous and thus does not require IRB approval.
    \item[] Guidelines:
    \begin{itemize}
        \item The answer \answerNA{} means that the paper does not involve crowdsourcing nor research with human subjects.
        \item Depending on the country in which research is conducted, IRB approval (or equivalent) may be required for any human subjects research. If you obtained IRB approval, you should clearly state this in the paper. 
        \item We recognize that the procedures for this may vary significantly between institutions and locations, and we expect authors to adhere to the NeurIPS Code of Ethics and the guidelines for their institution. 
        \item For initial submissions, do not include any information that would break anonymity (if applicable), such as the institution conducting the review.
    \end{itemize}

\item {\bf Declaration of LLM usage}
    \item[] Question: Does the paper describe the usage of LLMs if it is an important, original, or non-standard component of the core methods in this research? Note that if the LLM is used only for writing, editing, or formatting purposes and does \emph{not} impact the core methodology, scientific rigor, or originality of the research, declaration is not required.
    \item[] Answer: \answerYes{}
    \item[] Justification: We describe how LLMs are being used as part of the tools for evaluation data, as part of the research process as well as for generating data.
    \item[] Guidelines:
    \begin{itemize}
        \item The answer \answerNA{} means that the core method development in this research does not involve LLMs as any important, original, or non-standard components.
        \item Please refer to our LLM policy in the NeurIPS handbook for what should or should not be described.
    \end{itemize}

\end{enumerate}

\end{document}